\documentclass[10pt,twocolumn,letterpaper]{article}

\usepackage[pagenumbers]{cvpr} % To force page numbers, e.g. for an arXiv version

\usepackage[dvipsnames]{xcolor}

\newcommand{\todo}[1]{{\color{red}#1}}

\usepackage{colortbl}
\usepackage{multirow}
\usepackage{scrextend}
\usepackage{pythonhighlight}

\deffootnote{0em}{1.6em}{\thefootnotemark}

\definecolor{cvprblue}{rgb}{0.21,0.49,0.74}
\usepackage[pagebackref,breaklinks,colorlinks,citecolor=cvprblue]{hyperref}

\def\paperID{9719} % *** Enter the Paper ID here
\def\confName{CVPR}
\def\confYear{2024}

\title{Frozen Feature Augmentation for Few-Shot Image Classification}
\author{Andreas Bär\textsuperscript{1 2 *} \quad Neil Houlsby\textsuperscript{1} \quad Mostafa Dehghani\textsuperscript{1} \quad Manoj Kumar\textsuperscript{1 $\dagger$}\\
\textsuperscript{1}Google DeepMind \quad \textsuperscript{2}Technische Universität Braunschweig\\
{\tt\small andreas.baer@tu-braunschweig.de} \quad {\tt\small \{neilhoulsby, dehghani, mechcoder\}@google}
}

\begin{document}
\maketitle

\definecolor{tu0}{rgb}{0.7451, 0.1176, 0.2353}

\definecolor{tu1}{rgb}{1.0000, 0.8039, 0.0000}
\definecolor{tu11}{rgb}{1.0000, 0.8627, 0.3020}
\definecolor{tu12}{rgb}{1.0000, 0.9020, 0.4980}
\definecolor{tu13}{rgb}{1.0000, 0.9412, 0.6980}
\definecolor{tu14}{rgb}{1.0000, 0.9608, 0.8000}

\definecolor{tu2}{rgb}{0.9804, 0.4314, 0.0000}
\definecolor{tu21}{rgb}{0.9882, 0.6039, 0.3020}
\definecolor{tu22}{rgb}{0.9882, 0.7137, 0.4980}
\definecolor{tu23}{rgb}{0.9922, 0.8275, 0.6980}
\definecolor{tu24}{rgb}{0.9961, 0.8863, 0.8000}

\definecolor{tu3}{rgb}{0.6902, 0.0000, 0.2745}
\definecolor{tu31}{rgb}{0.7529, 0.2000, 0.4196}
\definecolor{tu32}{rgb}{0.8431, 0.4980, 0.6353}
\definecolor{tu33}{rgb}{0.9216, 0.7490, 0.8196}
\definecolor{tu34}{rgb}{0.9529, 0.8510, 0.8902}

\definecolor{tu4}{rgb}{0.4863, 0.8039, 0.9020}
\definecolor{tu41}{rgb}{0.6431, 0.8627, 0.9333}
\definecolor{tu42}{rgb}{0.7412, 0.9020, 0.9490}
\definecolor{tu43}{rgb}{0.8431, 0.9412, 0.9686}
\definecolor{tu44}{rgb}{0.8980, 0.9608, 0.9804}

\definecolor{tu5}{rgb}{0.0000, 0.5020, 0.7059}
\definecolor{tu51}{rgb}{0.3020, 0.6510, 0.7961}
\definecolor{tu52}{rgb}{0.5490, 0.7765, 0.8667}
\definecolor{tu53}{rgb}{0.7490, 0.8745, 0.9255}
\definecolor{tu54}{rgb}{0.8510, 0.9255, 0.9569}

\definecolor{tu6}{rgb}{0.0000, 0.3255, 0.4549}
\definecolor{tu61}{rgb}{0.2510, 0.4941, 0.5922}
\definecolor{tu62}{rgb}{0.5490, 0.6941, 0.7529}
\definecolor{tu63}{rgb}{0.7490, 0.8314, 0.8627}
\definecolor{tu64}{rgb}{0.8510, 0.8980, 0.9176}

\definecolor{tu7}{rgb}{0.0314, 0.0314, 0.0314}
\definecolor{tu71}{rgb}{0.3725, 0.3725, 0.3725}
\definecolor{tu72}{rgb}{0.5882, 0.5882, 0.5882}
\definecolor{tu73}{rgb}{0.7529, 0.7529, 0.7529}
\definecolor{tu74}{rgb}{0.8667, 0.8667, 0.8667}

\definecolor{tu8}{rgb}{0.7765, 0.9333, 0.0000}
\definecolor{tu81}{rgb}{0.8431, 0.9529, 0.3020}
\definecolor{tu82}{rgb}{0.8863, 0.9647, 0.4980}
\definecolor{tu83}{rgb}{0.9333, 0.9804, 0.6980}
\definecolor{tu84}{rgb}{0.9569, 0.9882, 0.8000}

\definecolor{tu9}{rgb}{0.5373, 0.6431, 0.0000}
\definecolor{tu91}{rgb}{0.6784, 0.7490, 0.3020}
\definecolor{tu92}{rgb}{0.7686, 0.8196, 0.4980}
\definecolor{tu93}{rgb}{0.8588, 0.8941, 0.6980}
\definecolor{tu94}{rgb}{0.9059, 0.9294, 0.8000}

\definecolor{tu10}{rgb}{0.0000, 0.4431, 0.3373}
\definecolor{tu101}{rgb}{0.3020, 0.6118, 0.5373}
\definecolor{tu102}{rgb}{0.5490, 0.7490, 0.7020}
\definecolor{tu103}{rgb}{0.7490, 0.8588, 0.8353}
\definecolor{tu104}{rgb}{0.8549, 0.9176, 0.9059}

\definecolor{tu110}{rgb}{0.8000, 0.0000, 0.6000}
\definecolor{tu111}{rgb}{0.8706, 0.3490, 0.7412}
\definecolor{tu112}{rgb}{0.9216, 0.6000, 0.8392}
\definecolor{tu113}{rgb}{0.9608, 0.8000, 0.9216}
\definecolor{tu114}{rgb}{0.9804, 0.8980, 0.9608}

\definecolor{tu120}{rgb}{0.4627, 0.0000, 0.4627}
\definecolor{tu121}{rgb}{0.5961, 0.2510, 0.5961}
\definecolor{tu122}{rgb}{0.7294, 0.4980, 0.7294}
\definecolor{tu123}{rgb}{0.8392, 0.6980, 0.8392}
\definecolor{tu124}{rgb}{0.9216, 0.8510, 0.9216}

\definecolor{tu130}{rgb}{0.4627, 0.0000, 0.3294}
\definecolor{tu131}{rgb}{0.6118, 0.3020, 0.5333}
\definecolor{tu132}{rgb}{0.7569, 0.5490, 0.6980}
\definecolor{tu133}{rgb}{0.8667, 0.7490, 0.8314}
\definecolor{tu134}{rgb}{0.9216, 0.8510, 0.9020}
\begin{abstract}

Vision foundation models are currently one of the main driving forces in computer vision research. Simply training a linear classifier or a lightweight model on top of model outputs or so-called `frozen features' leads to impressive performance on a number of tasks. Currently, frozen features are not modified during training of such lightweight models. On the other hand, when networks are trained directly on images, data augmentation is a standard recipe that improves performance with no additional overhead. In this paper, we conduct an extensive pilot study that explores applying data augmentations in the frozen feature space for few-shot image classification. We dub this type of augmentation `frozen feature augmentation (FroFA)'. Our study demonstrates that adopting deceptively simple pointwise FroFAs, such as brightness, can improve few-shot performance consistently across three network architectures, three large pretraining datasets, and eight transfer datasets.

% On few-shot ImageNet, we demonstrate that training a MAP Head with pointwise stylistic augmentation such as brightness or contrast, improves its performance consistently while geometric augmentations such as shear or rotation decrease performance. FroFa especially excels on smaller transfer datasets where it improves the mean 

% Third, a per-channel formulation (cFroFA) further increases the performance. Fourth, a simple sequential protocol of our best three (c)FroFA gives even stronger performance in some settings. Finally, we show that the results using a variation of brightness cFroFA are mostly consistent across three network architectures, two large pretraining datasets, and eight transfer datasets.

% In this paper, we extensively study this question in the few-shot setting.
%We chose the few-shot setting as we expect to observe significant effects on the performance by incorporating data augmentation in low-data regimes.
%To this end, we introduce a paradigm for reformulating eighteen standard image augmentations to frozen feature augmentation (FroFA).
%We found that reformulating geometric image augmentations to the FroFA paradigm do not improve performance, while stylistic ones have a strong impact on performance.
\end{abstract}

%\TODO{Decide whether to include or drop investigations on regularizations. We have numbers for label smoothing with the same setup (by EOB 16.11)}
\let\thefootnote\relax\footnotetext{\textsuperscript{*}Work conducted as Research Intern at Google DeepMind. \,\textsuperscript{$\dagger$}Project lead.}
\section{Introduction}
\label{sec:intro}

A prevalent trend now is to pretrain vision models on large datasets and adapt them downstream \cite{Zhai2022,Chen2023,oquab2023dinov2}. Notable, even training a simple linear layer or a light-weight model on top of vision transformer (ViT) outputs, also known as \textit{frozen features}, can yield remarkable performance across a number of diverse downstream tasks \cite{Radford2021,Dehghani2023,He2023}.

However, there is still an interest in training ViTs to achieve good performance on ImageNet-sized \cite{liu2021swin,wang2021pyramid} or smaller \cite{lee2021vision,liu2021efficient} datasets. In this setting, a crucial ingredient is data augmentation --- a predefined set of simple, stochastic input transformations. Simple but effective examples for image augmentations include random cropping which extracts a fixed-sized region from an image of arbitrary resolution, or pixel-wise modifications that change brightness, saturation, or contrast. These are complemented by more advanced augmentation strategies such as mixup \cite{Zhang2018h} or RandAugment \cite{Cubuk2020}.

\begin{figure}
  \centering
\includegraphics[width=\columnwidth]{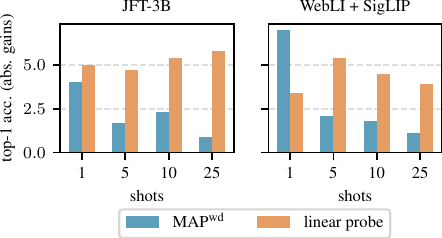}
  \caption{Few-shot results averaged across eight test sets, including ILSVRC-2012 \cite{Deng2009, Russakovsky2015}. We use cached features from an L/16 model \cite{Dosovitskiy2021} pretrained on JFT-3B \cite{Zhai2022} (left) or WebLI \cite{Chen2023} following a sigmoid language-image pretraining (SigLIP) \cite{Zhai2023} (right). Our method, \ie, a multi-head attention pooling \cite{Lee2019f} head trained with weight decay (MAP\textsuperscript{wd}) and frozen feature augmentation (FroFA), shows significant gains across all shots with respect to a weight-decayed MAP, \ie, MAP\textsuperscript{wd}, or an L2-regularized linear probe baseline, both without FroFA.}
  \label{fig:first_page}
\end{figure}

In this paper, we revisit standard image augmentation techniques in a data-constrained, few-shot frozen feature setting. In particular, we first stochastically transform frozen features and then train a lightweight model on top. Our only modification before applying image augmentations on top of frozen features is a point-wise scaling such that each feature value lies in $[0,1]$ or $[0, 255]$.

We investigate eighteen augmentations applied to frozen features extracted from vision transformers pretrained on JFT-3B \cite{Zhai2022}, ImageNet-21k \cite{Deng2009,Russakovsky2015}, or WebLI \cite{Chen2023}. We train a small lightweight multi-head attention pooling (MAP) \cite{Lee2019f,Zhai2022} head using these augmented inputs and evaluate its performance across eight downstream image classification datasets, where we on average achieve significant gains (see \cref{fig:first_page}). Our major insights are as follows:

\begin{enumerate}
    \item Geometric augmentations that modify the shape and structure of two-dimensional frozen features always lead to worse performance on ImageNet. On the other hand, simple stylistic (point-wise) augmentations, such as brightness, contrast, and posterize, give steady improvements on 1-, 5-, and 10-shot settings.
    \item Unlike traditional image augmentations that apply a single randomly sampled value across the entire image, we introduce per-channel stochasticity by sampling independent random values for each channel. For example, on the 5-shot setting, we improve accuracy over a well-tuned MAP and linear probe baseline by 0.5\% absolute and 0.8\% absolute, respectively.
    \item While FroFA provides modest but significant improvements on ImageNet, it excels on smaller transfer datasets. Across seven downstream datasets, FroFA outperforms the mean accuracy of the MAP baseline in the 5 shot setting by 3.2\% absolute and the linear probe baseline by 4.2\% absolute.
\end{enumerate}
% Formulate the research question by connecting feature caching and data augmentation
% In this paper, we design a framework for frozen feature augmentation (FroFA) and perform a thorough investigation.
% % Motivate few-shot setting as we hope to see large gains here.
% We focus on few-shot image classification as we expect to observe significant effects using data augmentation in low-data regimes.
% List contributions and highlight the most important ones
% . Our insights are: First, simple stylistic image augmentations reinterpreted as FroFAs, \eg, brightness, significantly improve performance across 1- to 25-shot settings --- the gains decrease with higher shots. Second, a per-channel FroFA formulation in some cases further boosts performance. Third, a first step into sequential FroFA protocols shows a positive outcome and opens up the path for future works in this direction. Fourth, our results transfer with regard to architecture, pretraining and transfer datasets. Finally, while linear probe remains a strong baseline in the 1-shot setting, we are able to surpass it in most 5- to 25-shot settings (\cf \cref{fig:first_page}).

% The rest of the paper is organized as follows: We discuss related works in \cref{sec:related}. Next, we introdcue our feature caching and data augmentation framework in \cref{sec:framework}. We follow up by the experimental setup in \cref{sec:exp_setup} and the experimental results in \cref{sec:exp_results}. Finally, we conclude in \cref{sec:conclusions}.
\section{Related Works}
\label{sec:related}

\textbf{Transfer learning on few-shot data}: State-of-the-art vision models  \cite{Dosovitskiy2021,Zhai2022,Chen2023,Dehghani2023} are typically pretrained on large-scale datasets, \eg, ImageNet-21k \cite{Deng2009,Russakovsky2015} or versions of JFT \cite{Hinton2014,Zhai2022}, before transferred to other middle-scale to small-scale ones, \eg, CIFAR10 \cite{Krizhevsky2009}, ILSVRC-2012 \cite{Deng2009,Russakovsky2015}, or SUN397 \cite{Xiao2010,Xiao2016}. Depending on the model size, efficient transfer learning becomes a challenge. Many methods have been proposed for large language models (LLMs), \eg, adapters \cite{Houlsby2019}, low-rank adaptation (LoRA) \cite{Hu2022}, or prompt tuning \cite{Lester2021}, of which some have been successfully adapted to computer vision \cite{Zhang2022a,Chen2022,Jia2022,Gao2023}.
CLIP-Adapter \cite{Gao2023} builds on the power of contrastive language-image pretraining (CLIP) \cite{Radford2021} and combines it with adapters \cite{Houlsby2019}.
A follow-up work \cite{Zhang2022a} proposes TiP-Adapter which uses a query-key cache model \cite{Grave2017,Orhan2018} instead of a gradient descent approach.
Inspired by the success of prompt tuning in LLMs \cite{Lester2021}, Jia \etal propose visual prompt tuning at the model input \cite{Jia2022}. On the other hand, AdaptFormer \cite{Chen2022} uses additional intermediate trainable layers to finetune a frozen vision transformer \cite{Dosovitskiy2021}.

In contrast, we do not introduce additional prompts \cite{Jia2022} or intermediate parameters \cite{Gao2023,Chen2022} that require backpropagating through the network. Instead, we train a small network on top of frozen features coming from a vision transformer. This aligns with linear probing \cite{Radford2021} which is typically used to transfer vision models to other tasks \cite{He2023,Zhai2022,Dehghani2023} --- our objective.

In addition, we focus our experiments around transfer learning on few-shot data \cite{Vinyals2016,Lake2019}. Although not surprising, few-shot results obtained by Dehghani \etal \cite{Dehghani2023} clearly show significant gaps between linear probing and full finetuning. We take these results as an incentive to improve upon linear probing.

\textbf{Data augmentation}: One go-to method to improve performance while training in a low-data regime is data augmentation \cite{Steiner2022}. Some prominent candidates in computer vision are AutoAugment \cite{Cubuk2019}, AugMix \cite{Hendrycks2020}, RandAugment \cite{Cubuk2019}, and TrivialAugment \cite{Mueller2021}. These methods typically combine low-level image augmentations together to augment the input. Although some works propose augmentations in feature space \cite{DeVries2017,Verma2019,Liu2018a,Liu2023a,Kumar2019a}, a large-scale empirical study on \textit{frozen features} of single-modal vision models does not exist.

To this end, we investigate frozen feature augmentation (FroFA) by reformulating eighteen image augmentations. In particular, we consider a subset used in AutoAugment \cite{Cubuk2019}, inception crop \cite{Szegedy2016}, mixup \cite{Zhang2018h,Verma2019}, and the recently introduced patch dropout \cite{Liu2023}.
\section{Framework Overview}
\label{sec:framework}
In this section, we give an overview of our framework.
%\cref{ssec:notation} introduces the notation followed by our caching and training pipeline in \cref{sec:train_cf}. Finally, we describe our proposed frozen feature augmentation (FroFA) in \cref{sec:frofa}.

\subsection{Notation}
\label{ssec:notation}
Let $\boldsymbol{x}\in\mathbb{I}^{H\times W\times 3}$ be an RGB image  of height $H$, width $W$, and $\mathbb{I}=[0, 1]$. A classification model processes $\boldsymbol{x}$ and outputs class scores $\boldsymbol{y}\in[0,1]^{S}$ for each class in a predefined set of classes $\mathcal{S}$, with $S=|\mathcal{S}|$. Let $L$ and $D$ be the number of intermediate layers and the number of features of a multi-layer classification model, respectively. We describe the intermediate feature representations of $\boldsymbol{x}$ as $\boldsymbol{f}=\boldsymbol{f}^{(\ell)}=(f^{(\ell)}_{d})\in\mathbb{R}^{D}$, with layer index $\ell\in\lbrace 1, ..., L\rbrace$ and feature index $d\in\lbrace 1, ..., D\rbrace$. In the vision transformer \cite{Kolesnikov2020} architecture, $\boldsymbol{f}=\boldsymbol{f}^{(\ell)}=(f^{(\ell)}_{n,c})\in\mathbb{R}^{N\times C}$ is a two-dimensional entity, where $N$ and $C$ are the number of patches and number of per-patch channels, respectively. In addition, we introduce the patch index $n\in\lbrace 1, ..., N \rbrace$ and the per-patch channel index $c\in\lbrace 1, ..., C \rbrace$.
% \begin{equation}
%   E = m\cdot c^2
%   \label{eq:important}
% \end{equation}
% and
% \begin{equation}
%   v = a\cdot t.
%   \label{eq:also-important}
% \end{equation}

\subsection{Training on Cached Features}
\label{sec:train_cf}
\begin{figure}
  \centering
  \begin{subfigure}{\columnwidth}
%trim={left bottom right top}
\includegraphics[trim={0 1cm 0 0},clip,width=\columnwidth]{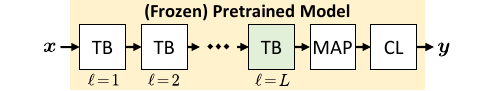}
    \caption{Step 1: Select a (frozen) pretrained model and a layer for caching.}
    \label{fig:framework_overview-a}
  \end{subfigure}
  \hfill
  \begin{subfigure}{\columnwidth}
    \includegraphics[trim={0 0.4cm 0 -0.2cm},clip,width=\columnwidth]{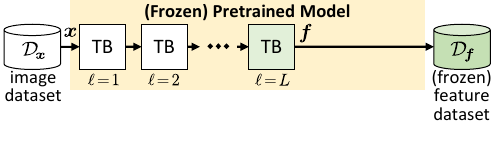}
    \caption{Step 2: Process an image dataset and cache the (frozen) features.}
    \label{fig:framework_overview-b}
  \end{subfigure}
   \begin{subfigure}{\columnwidth}
    \includegraphics[trim={0 0.4cm 0 -0.2cm},clip,width=\columnwidth]{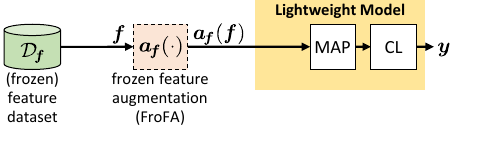}
    \caption{Step 3: Train on (augmented) frozen features.}
    \label{fig:framework_overview-c}
  \end{subfigure}
  \caption{\textbf{Pipeline for caching and training on (frozen) features}. (\ref{fig:framework_overview-a}): Given a (frozen) pretrained vision transformer, with $L$ Transformer blocks (TBs), a multi-head attention pooling (MAP) layer, and a classification layer (CL), we select its $L$-th Transformer block for caching. (\ref{fig:framework_overview-b}): Next, we feed images $\boldsymbol{x}\in\mathcal{D}_{\boldsymbol{x}}$ to cache (frozen) features $\boldsymbol{f}\in\mathcal{D}_{\boldsymbol{f}}$. (\ref{fig:framework_overview-c}): Finally, we use $\mathcal{D}_{\boldsymbol{f}}$ to train a lightweight model on top. We investigate frozen feature augmentation (FroFA) $\boldsymbol{a}_{\boldsymbol{f}}\in\mathcal{A}_{\boldsymbol{f}}$ in this scenario.}
  \label{fig:framework_overview}
\end{figure}

We investigate pretrained vision transformers \cite{Kolesnikov2020} with $L$ transformer blocks (TBs) followed by a multi-head attention pooling (MAP) \cite{Lee2019f} and a classification layer (CL). \cref{fig:framework_overview-a} presents a simplified illustration. For simplicity, we neglect all operations before the first transformer block (e.g., patchifying, positional embedding, etc.). 

To cache intermediate feature representations, we process each image $\boldsymbol{x}$ from an image dataset $\mathcal{D}_{\boldsymbol{x}}$ through the network up until transformer block $L$. Next, we store the resulting features $\boldsymbol{f}$. After processing $\mathcal{D}_{\boldsymbol{x}}$ we obtain a (frozen) feature dataset $\mathcal{D}_{\boldsymbol{f}}$, with $\boldsymbol{f}\in\mathcal{D}_{\boldsymbol{f}}$  (\cref{fig:framework_overview-b}).

Finally, we train a lightweight model using the cached (frozen) features. \cref{fig:framework_overview-c} shows an example where a single MAP layer followed by a classification layer is trained using the feature dataset $\mathcal{D}_{\boldsymbol{f}}$. Since our focus is fast training, we defer a detailed analysis on larger models to future work.

\subsection{Frozen Feature Augmentation (FroFA)}\label{sec:frofa}

Data augmentation is a common tool to improve generalization and is typically applied on the input, or in our case: images. A natural question arises: How to map such image augmentations to intermediate feature representations?

Recall that the feature representation $\boldsymbol{f}=(f_{n,c})\in\mathbb{R}^{N\times C}$ (layer index $\ell$ omitted) is two-dimensional. We first reshape it to a three-dimensional representation, \ie,
\begin{equation}\label{eq:reshaped_features}
\boldsymbol{f}^*=(f^*_{n_1,n_2,c})\in\mathbb{R}^{\sqrt{N}\times\sqrt{N}\times C}.
\end{equation}
We further define 
\begin{equation}\label{eq:single_feature}
\boldsymbol{f}^*_{c}=\boldsymbol{f}^*_{:,:,c}\in\mathbb{R}^{\sqrt{N}\times\sqrt{N}\times 1}
\end{equation}
as a two-dimensional representation of the $c$-th channel.
%\cref{fig:image-feature-pair} shows an image-feature pair, where we illustrate the first six reshaped channels. We observe that some channels may have a similar structure to the original image ($\boldsymbol{f}^*_{c=6}$, highlighted in \textcolor{red}{red}). This motivates the use of image augmentations in the feature space.

Images and feature representations differ in two fundamental aspects: channel dimensionality and value range. Before adapting image augmentations to the feature space, it is crucial to handle these differences.

\textbf{Channel dimensionality}:
RGB images have just three channels while intermediate representations possess an arbitrary number of channels.  To address this, we ignore image augmentations that rely on three color channels, \eg, color jitter, and consider augmentations which can have an arbitrary number of channels instead, denoted as $C_{\boldsymbol{a}}$, covering a majority of commonly applied image augmentations.

\textbf{Value range}: RGB values lie within a specific range $\mathbb{I}$, \eg, $\mathbb{I}=[0, 1]$ or $\mathbb{I}=\lbrace0, ..., 255\rbrace\subset\mathbb{N}_0$, while in theory features have no such constraints. 
Assuming $H=\sqrt{N}$ and $W=\sqrt{N}$, we define an image augmentation as
\begin{equation}\label{eq:image_aug}
  \boldsymbol{a}_{\boldsymbol{x}}:\mathbb{I}^{\sqrt{N}\times \sqrt{N}\times C_{\boldsymbol{a}}} \to \mathbb{I}^{\sqrt{N}\times \sqrt{N}\times C_{\boldsymbol{a}}}, \boldsymbol{a}_{\boldsymbol{x}}\in\mathcal{A}_{\boldsymbol{x}},
\end{equation}
where $\mathcal{A}_{\boldsymbol{x}}$ is the set of image augmentations and $C_{\boldsymbol{a}}=C$ is an arbitrary number of channels.
To also address the value range mismatch, we introduce a deterministic feature-to-image mapping
\begin{equation}\label{eq:f_to_x}
\boldsymbol{t}_{\boldsymbol{f}\rightarrow\boldsymbol{x}}: \mathbb{R}^{\sqrt{N}\times \sqrt{N}\times C_{\boldsymbol{t}}} \to \mathbb{I}^{\sqrt{N}\times \sqrt{N}\times C_{\boldsymbol{t}}}
\end{equation}
that maps each element of $\boldsymbol{f}^*$ \eqref{eq:reshaped_features} from $\mathbb{R}$ to $\mathbb{I}$.
In our experiments, we use 
\begin{equation}\label{eq:f_to_x_specific}
\boldsymbol{x}_{\boldsymbol{f}}=\boldsymbol{t}_{\boldsymbol{f}\rightarrow\boldsymbol{x}}(\boldsymbol{f}^*)=\frac{\boldsymbol{f}^* - f_{\mathrm{min}}}{f_{\mathrm{max}} - f_{\mathrm{min}}},
\end{equation}
where $f_{\mathrm{min}}$ and $f_{\mathrm{max}}$ are the minimum and maximum value of $\boldsymbol{f}^*$, respectively, with elements of $\boldsymbol{x}_{\boldsymbol{f}}$ now in $\mathbb{I}=[0, 1]$. 
We further define an image-to-feature mapping
\begin{equation}\label{eq:x_to_f}
\boldsymbol{t}_{\boldsymbol{f}\leftarrow\boldsymbol{x}}: \mathbb{I}^{\sqrt{N}\times \sqrt{N}\times C_{\boldsymbol{t}}}\to\mathbb{R}^{\sqrt{N}\times \sqrt{N}\times C_{\boldsymbol{t}}}
\end{equation}
that maps $\boldsymbol{x}_{\boldsymbol{f}}$ back to the original feature value range, with $C_{\boldsymbol{t}}=C$ by default.
In this case, we simply invert \eqref{eq:f_to_x} and use
\begin{equation}\label{eq:x_to_f_specific}
\boldsymbol{f}^*=\boldsymbol{t}_{\boldsymbol{f}\leftarrow\boldsymbol{x}}(\boldsymbol{x}_{\boldsymbol{f}})=\boldsymbol{x}_{\boldsymbol{f}} \cdot (f_{\mathrm{max}} - f_{\mathrm{min}}) + f_{\mathrm{min}}.
\end{equation}
Combining \eqref{eq:image_aug}, \eqref{eq:f_to_x}, and \eqref{eq:x_to_f}, we obtain a generic (frozen) feature augmentation (FroFA) as a function composition 
\begin{equation}\label{eq:feature_aug}
\boldsymbol{a}_{\boldsymbol{f}}=\boldsymbol{t}_{\boldsymbol{f}\leftarrow\boldsymbol{x}}\circ \boldsymbol{a}_{\boldsymbol{x}}\circ\boldsymbol{t}_{\boldsymbol{f}\rightarrow\boldsymbol{x}}.
\end{equation}
% In cases where image-to-feature/feature-to-image mapping is not necessary, we simply omit it. (I removed it because it is not clear what cases this is not necessary)

We use three variations of $\boldsymbol{a}_{\boldsymbol{f}}$:

\begin{enumerate}
    \item \textbf{(Default) FroFA}: We apply $\boldsymbol{a}_{\boldsymbol{f}}$ \eqref{eq:feature_aug} once across the entire feature representation. We set $C_{\boldsymbol{a}}=C_{\boldsymbol{t}}=C$ and compute $f_{\mathrm{min}}$ and $f_{\mathrm{max}}$ in \eqref{eq:f_to_x_specific}, \eqref{eq:x_to_f_specific} across all elements of $\boldsymbol{f}^*$. Further, as normally done in pixel space, $\boldsymbol{a}_{\boldsymbol{x}}$ \eqref{eq:image_aug} samples a random augmentation value and changes all elements of $\boldsymbol{x}_{\boldsymbol{f}}$ using the same value. For example, employing random contrast in a FroFA fashion scales each element of $\boldsymbol{x}_{\boldsymbol{f}}$ by the \textit{exact same randomly sampled factor}. 
    \item \textbf{Channel FroFA (cFroFA)}:
    For each channel in the mapped features $\boldsymbol{x}_{\boldsymbol{f}}$ \eqref{eq:f_to_x_specific}, $\boldsymbol{a}_{\boldsymbol{x}}$ \eqref{eq:image_aug} samples a random augmentation value \textit{per channel} and applies that value to all elements in that channel. By using cFroFA for our random contrast example, we obtain \textit{$C$ independently sampled scaling factors, one for each channel}. 
    \item \textbf{Channel$^2$ FroFA (c$^2$FroFA)}: In addition to applying augmentations per channel as done in cFroFA, $\boldsymbol{t}_{\boldsymbol{f}\rightarrow\boldsymbol{x}}$ \eqref{eq:f_to_x} and $\boldsymbol{t}_{\boldsymbol{x}\leftarrow\boldsymbol{f}}$ \eqref{eq:x_to_f} also operate per channel. In this case,  $f_\mathrm{min}$ and $f_\mathrm{max}$ are the per-channel maximum and minimum, respectively. In contrast, FroFA and cFroFA use the maximum and minimum across the entire feature. We denote this variant as c$^2$FroFA since both the mappings \eqref{eq:f_to_x}, \eqref{eq:x_to_f} and the augmentation \eqref{eq:image_aug} are applied on a per-channel basis. Although not adding additional stochasticity, we found that for random brightness this variant gives more stable results across a range of augmentation hyper parameters.
\end{enumerate}
While an element-wise FroFA might seem like a natural next step, our initial experiments lead to significantly worse results. We hypothesize that per-element augmentations might lead to substantial changes in the feature appearance.
\section{Experimental Setup}
\label{sec:exp_setup}
In this section, we introduce our experimental setup.

\subsection{Network Architectures}
We employ the following pretrained vision transformers from prior work: Ti/16 \cite{Touvron2021}, B/16 \cite{Dosovitskiy2021}, and L/16 \cite{Dosovitskiy2021}. Further, we follow \cite{Zhai2022} and employ a lightweight multi-head attention pooling (MAP) layer \cite{Lee2019f} before the final classification layer on top of the frozen features (\cf \cref{sec:frofa}).

\subsection{Datasets}\label{sec:dataset_details}
\textbf{Pretraining}: We consider three datasets: JFT-3B, ImageNet-21k, and WebLI. First introduced by Hinton \etal \cite{Hinton2014}, JFT is now a widely used proprietary, large-scale dataset \cite{Chollet2017,Sun2017,Kolesnikov2020,Dosovitskiy2021,Dai2021,Zhai2022,Kossen2023,Chen2023}. For our investigations we use the JFT-3B version following Zhai \etal \cite{Zhai2022}. It consists of nearly 3 billion multi-labeled images following a class-hierarchy of 29,593 labels. We further use ImageNet-21k \cite{Deng2009, Russakovsky2015} which consists of 14,197,122 (multi)-labeled images and 21,841 distinct labels. We equally split the first 51,200 images into a validation and test set and use the remaining 14,145,922 images for training. As a third dataset, we use WebLI \cite{Chen2023} which is a recently introduced web-scale multilingual image-text dataset. Please refer to the Appendix, \cref{sec:appendix_datasets}, for more details.

\textbf{Few-shot transfer}: After pretraining we use eight datasets for few-shot transfer: ILSVRC-2012 \cite{Deng2009, Russakovsky2015},
CIFAR10 \cite{Krizhevsky2009}, CIFAR100 \cite{Krizhevsky2009}, DMLab \cite{Beattie2016,Zhai2020}, DTD \cite{Cimpoi2014}, Resisc45 \cite{Cheng2017}, SUN397 \cite{Xiao2010,Xiao2016}, and SVHN \cite{Netzer2011}.

ILSVRC-2012, also known as ImageNet-1k, is a slimmed version of ImageNet-21k and contains 1,281,167 training images of 1,000 classes. We use it as our main few-shot benchmark throughout the paper. We randomly sample 1-shot, 5-shot, 10-shot, and 25-shot versions from the first 10\% of the training set. We further create additional disjoint sets by using the next four 10\% fractions of the training set. In addition, we follow previous works \cite{Beyer2022} and create a `minival' set using the last 1\% (12,811 images) of the ILSVRC-2012 training set. The `minival' set is used for hyper parameter tuning and design decisions while the official ILSVRC-2012 validation set is used as a test set. In summary, our setup consists of 1,000, 5,000, 10,000, or 25,000 training images, 12,811 validation images (`minival'), and 50,000 test images (`validation').

For the other seven datasets, we also select a training, validation, and test split and create few-shot versions.  More details on how these splits are created can be found in the Appendix, \cref{sec:appendix_datasets}. We follow a similar procedure as with ILSVRC-2012 and use 10\% of the training images to create 1-shot, 5-shot, 10-shot, and 25-shot versions of each dataset. We further use each validation set for hyper parameter tuning and report final results on the respective test set.

%CIFAR10 is a popular image classification dataset and contains 60,000 natural images of objects categorized into 10 evenly split classes. CIFAR100 extends CIFAR10 to 60,000 images and 100 evenly split classes. DMLab contains six different categories of 110,913 frames from a rendered 3D maze. DTD comprises 5,640 images of 47 evenly split and distinguishable textural patterns. Resisc45 is an image scene classification dataset with 31,500 images of 45 classes. SUN397 is another scene classification dataset containing 108,753 images divided into 397 categories. Finally, SVHN contains 99,289 training and test images of 10 house numbers. We select a training, validation, and test split for each transfer dataset and create few-shot versions. More details on how these splits are created can be found in the Appendix, \cref{sec:appendix_datasets}.

\subsection{Data Augmentation}\label{sec:augmentation_details}
We reuse the set of augmentations first defined in AutoAugment \cite{Cubuk2019} and adopted in later works, such as RandAugment \cite{Cubuk2020} and TrivialAugment \cite{Mueller2021}. In addition, we also consider a few other image augmentations \cite{Szegedy2016,Zhang2018h,Liu2023}. We select \textit{five geometric} augmentations, \ie, {rotate}, {shear-x}, {shear-y}, {translate-x}, and {translate-y}; \textit{four crop \& drop} augmentations, \ie, {crop}, {resized crop}, {inception crop} \cite{Szegedy2016}, and {patch dropout} \cite{Liu2023}; \textit{seven stylistic} augmentations, \ie, {brightness}, {contrast}, {equalize}, {invert}, {posterize}, {sharpness}, and {solarize}; and \textit{two other} augmentations, \ie, {JPEG} and {mixup} \cite{Zhang2018h}. In total, we end up with \textit{eighteen distinct augmentations}. Note that all data augmentations incorporate random operations, \eg, a random shift in x- and y-direction (translate-x and translate-y, respectively), a randomly selected set of patches (patch dropout), a random additive value to each feature (brightness), or a random mix of two features and their respective classes (mixup). Please refer to the Appendix, \cref{sec:appendix_augmentation}, for more details.

We focus on the following set of experiments:
\begin{enumerate}
    \item We investigate FroFA for all eighteen augmentations.
    \item  For our top-performing FroFAs, namely, brightness, contrast, and posterize, we incorporate additional stochasticity using cFroFA and c$^2$FroFA variants (\cf \cref{sec:frofa}).
    \item We investigate a sequential protocol where two of the best three (c/c$^2$)FroFAs are arranged sequentially, namely, brightness c$^2$FroFA, contrast FroFA, and posterize cFroFA. We test all six possible combinations.
    \item Finally, we also apply variations of RandAugment \cite{Cubuk2020} and TrivialAugment \cite{Mueller2021} directly on top of cached frozen features. More details and results can be found in the Appendix, \cref{sec:appendix_augmentation,sec:appendix_exp_results}, respectively.
\end{enumerate}

\subsection{Training \& Evaluation Details}\label{sec:training_details}
We describe some base settings for pretraining, few-shot learning, and evaluation. Please refer to Appendix, \cref{sec:appendix_training_details} for more training details.

\textbf{Pretraining}: We use the \texttt{Big Vision}\footnote{\url{https://github.com/google-research/big_vision}} code base for pretraining. We take the Ti/16, B/16, and L/16 models pretrained on JFT-3B from Zhai \etal \cite{Zhai2022}. In addition, we pretrain Ti/16, B/16 and L/16 on ImageNet-21k following the settings
of Steiner \etal \cite{Steiner2022}. To further explore transfer capabilities we also use an L/16 model with sigmoid language-image pretraining (SigLIP) \cite{Zhai2023} on WebLI \cite{Chen2023}.

\textbf{Few-shot learning}:
We use the \texttt{Scenic}\footnote{\url{https://github.com/google-research/scenic}} code base \cite{Dehghani2022} for few-shot learning. We train the lightweight MAP-based head by sweeping across five batch sizes (32, 64, 128, 256, and 512), four learning rates (0.01, 0.03, 0.06, and 0.1), and five training step sizes (1,000; 2000; 4,000; 8,000; and 16,000), yielding 100 configurations for each shot.
%In total, we obtain 100 configurations for each shot.
We use the respective validation set for early stopping and to find the best sweep setting. Our cached-feature setup fits on a single-host TPUv2 platform where our experiments run in the order of minutes.

\textbf{Evaluation}: We report the top-1 accuracy across all our few-shot datasets. On ILSVRC-2012, we tune few-shot models exclusively on our validation set (our ILSVRC-2012 `minival', \cf \cref{sec:dataset_details}) and report results on our test set (official ILSVRC-2012 `validation' set, \cf \cref{sec:dataset_details}).
% We follow the same convention for CIFAR100, Resisc45, and SUN397.

\subsection{Baseline Models}\label{sec:baselines}
We establish two baselines: MAP and linear probe.

\textbf{MAP}: We first cache the $N\!\times\!C$-shaped (frozen) features from the last transformer block. Afterwards, we train a lightweight MAP head from scratch using the cached features followed by the final classification layer (\cf \cref{fig:framework_overview}). For simplicity, the MAP head follows the same architectural design as the underlying pretrained model. In some experiments, we additionally apply weight decay (wd), denoted as MAP\textsuperscript{wd}. We sweep across \todo{[ADD VALUES]} and use the respective validation set for early stopping and to find the best sweep setting.

\textbf{Linear probe}: We use cached $1\!\times\!C$-shaped outputs from the pretrained MAP head to solve an L2-regularized regression problem with a closed-form solution \cite{Zhai2022}. We sweep the L2 decay factor using exponents of 2 ranging from -20 up to 10. This setting is our auxiliary baseline.
% \begin{table}
%   \centering
%   \begin{tabular}{@{}lc@{}}
%     \toprule
%     Method & Frobnability \\
%     \midrule
%     Theirs & Frumpy \\
%     Yours & Frobbly \\
%     Ours & Makes one's heart Frob\\
%     \bottomrule
%   \end{tabular}
%   \caption{Results.   Ours is better.}
%   \label{tab:example}
% \end{table}

\section{Finding the Optimal FroFA Setup}
\label{sec:exp_results}

We focus our first investigations on an L/16 model pretrained on JFT-3B, \ie, our largest model and largest image classification pretraining dataset, followed by few-shot learning on subsets of ILSVRC-2012 training set, \ie, our largest few-shot dataset. We will refer to this setup as \textit{our L/16 JFT-3B base setup}.

\subsection{Baseline Performance}
We first report the baseline performance in \cref{tab:baselines}.
We observe a large gap between MAP and linear probe in the 1-shot setting (-8.6\% absolute) which significantly decreases in the 5-, 10-, and 25-shot settings to -0.8\%, -0.6\%, and +0.8\% absolute, respectively.
%We attribute the large gap in the 1-shot setting to the L2 regularization in the linear probe baseline.

In the following, our main point of comparison is the MAP baseline. This might be counter-intuitive since the performance is worse than linear probe in most cases. 
However, the higher input dimensionality in the MAP-based setting (\cf \cref{sec:baselines}) gives us the option to reshape the input to three dimensions (\cf \cref{sec:frofa}) which opens up more room and variety for frozen feature augmentations (FroFAs). Later in \cref{ssec:linear_probe_comparison}, we compare the performance of our best augmentations to the linear probe baseline.

% % Double digit with standard errors
% \begin{table}
% \setlength{\tabcolsep}{2.8pt}
% \footnotesize
%   \centering
%   \begin{tabular}{@{}lcccc@{}}
%     \toprule
%     Method & 1-shot & 5-shot & 10-shot& 25-shot \\
%     \midrule
%     MAP &  \cellcolor{gray!20}57.93±0.25&\cellcolor{gray!20}78.77±0.12&\cellcolor{gray!20}80.91±0.02&\cellcolor{gray!60}{83.19}±0.04\\
%     Linear probe &  \cellcolor{gray!60}{66.51}±0.38 & \cellcolor{gray!60}{79.60}±0.06 & \cellcolor{gray!60}81.45±0.07 & \cellcolor{gray!20}82.43±0.08\\
%     %MAP (full) &  \cellcolor{gray!20}{64.49}±0.15 & \cellcolor{gray!60}{80.41}±0.10 & \cellcolor{gray!60}{82.08}±0.09 & \cellcolor{gray!60}{83.58}±0.06\\
%     \bottomrule
%   \end{tabular}
%   \caption{\textbf{Baseline top-1 accuracy} with standard errors on \textit{our} ILSVRC-2012 test set. 
%   We use the L/16 JFT-3B base setup (\cf \cref{sec:exp_results}) and follow the respective baseline setting (\cf \cref{sec:baselines}). The best setting for each baseline is found using \textit{our} ILSVRC-2012 validation set. Further, each shot is sampled five times. The \colorbox{gray!60}{best} and \colorbox{gray!20}{second-best} results per shot are highlighted.}
%   \label{tab:baselines}
% \end{table}
% Single digit and no standard errors
\begin{table}
\setlength{\tabcolsep}{2.8pt}
\footnotesize
  \centering
  \begin{tabular}{@{}lcccc@{}}
    \toprule
    Method & 1-shot & 5-shot & 10-shot& 25-shot \\
    \midrule
    MAP & 57.9&78.8&80.9&\textbf{83.2}\\
    Linear probe &  \textbf{66.5} & \textbf{79.6} & \textbf{81.5} & 82.4\\
    \bottomrule
  \end{tabular}
  \caption{\textbf{Average top-1 accuracy for baseline settings} on \textit{our} ILSVRC-2012 test set. 
  We use the L/16 JFT-3B base setup (\cf \cref{sec:exp_results}) and follow the respective baseline setting (\cf \cref{sec:baselines}). The best setting for each baseline is found using \textit{our} ILSVRC-2012 validation set. Further, each shot is sampled five times. The best result per shot is boldfaced.}
  \label{tab:baselines}
\end{table}

\subsection{Default FroFA}
\begin{table*}
\setlength{\tabcolsep}{2.8pt}
\footnotesize
  \centering
  \begin{tabular}{@{}lccccccccccccccccccccccc@{}}
    \toprule
    &&&\multicolumn{5}{c}{Geometric}&&\multicolumn{4}{c}{Crop \& drop}&&\multicolumn{7}{c}{Stylistic}&&\multicolumn{2}{c}{Other}\\
    Shots & MAP && \rotatebox{90}{{rotate}}& \rotatebox{90}{{shear-x}}&\rotatebox{90}{{shear-y}}&\rotatebox{90}{{translate-x}}&\rotatebox{90}{{translate-y}}&&\rotatebox{90}{{crop}}&\rotatebox{90}{{res.\ crop}}&\rotatebox{90}{{incept.\ crop}}&\rotatebox{90}{{patch drop.}}&&\rotatebox{90}{{\textbf{brightness}}}&\rotatebox{90}{{\textbf{contrast}}}&\rotatebox{90}{{equalize}}&\rotatebox{90}{{invert}}&\rotatebox{90}{{\textbf{posterize}}}&\rotatebox{90}{{sharpness}*}&\rotatebox{90}{{solarize}*}&&\rotatebox{90}{{JPEG}*}&\rotatebox{90}{{mixup}}\\
    \midrule
    %\cellcolor{tu0!60}%\cellcolor{tu0!40}
    %\cellcolor{tu9!60}\cellcolor{tu9!40}\cellcolor{tu9!20}
    1 &  57.9&& \cellcolor{tu0!60}$-$1.3 & \cellcolor{tu0!40}$-$0.6 & \cellcolor{tu0!40}$-$0.8 &\cellcolor{tu0!60}$-$1.2&\cellcolor{tu0!60}$-$1.4&&\cellcolor{tu9!80}$+$3.0&\cellcolor{tu9!40}$+$1.9&$+$0.0&\cellcolor{tu9!20}$+$0.4&&\cellcolor{tu9!}$+$4.8&\cellcolor{tu9!60}$+$2.8&\cellcolor{tu9!40}$+$1.0&\cellcolor{tu9!60}$+$2.7&\cellcolor{tu9!80}$+$3.7&\cellcolor{tu0!40}$-$0.1&\cellcolor{tu9!40}$+$1.0&&\cellcolor{tu0!40}$-$0.1&\cellcolor{tu0!60}$-$1.4\\
    5 &  78.8 &&\cellcolor{tu0!40}$-$0.3 & \cellcolor{tu0!40}$-$0.2 & \cellcolor{tu0!40}$-$0.2 &\cellcolor{tu0!40}$-$0.3&\cellcolor{tu0!40}$-$0.3&&$+$0.0&\cellcolor{tu0!40}$-$0.2&$+$0.0&$+$0.0&&\cellcolor{tu9!40}$+$1.1&\cellcolor{tu9!20}$+$0.8&\cellcolor{tu9!20}$+$0.5&\cellcolor{tu0!40}$-$0.3&\cellcolor{tu9!20}$+$0.8&\cellcolor{tu9!20}$+$0.1&\cellcolor{tu0!40}$-$0.1&&\cellcolor{tu0!40}$-$0.3&\cellcolor{tu0!40}$-$0.3\\
    10 &  80.9 &&\cellcolor{tu0!40}$-$0.2 & \cellcolor{tu0!40}$-$0.1 & \cellcolor{tu0!40}$-$0.1 &\cellcolor{tu0!40}$-$0.2&\cellcolor{tu0!40}$-$0.2&&$+$0.0&\cellcolor{tu0!40}$-$0.2&$+$0.0&$+$0.0&&\cellcolor{tu9!20}$+$0.6&\cellcolor{tu9!20}$+$0.6&\cellcolor{tu9!20}$+$0.4&$+$0.0&\cellcolor{tu9!20}$+$0.6&\cellcolor{tu9!20}$+$0.1&$+$0.0&&\cellcolor{tu0!40}$-$0.1&\cellcolor{tu9!20}$+$0.2\\
    25 &  83.2 && \cellcolor{tu0!40}$-$0.2 & \cellcolor{tu0!40}$-$0.1 & \cellcolor{tu0!40}$-$0.2&\cellcolor{tu0!40}$-$0.1&\cellcolor{tu0!40}$-$0.2&&{$+$0.0}&\cellcolor{tu0!40}$-$0.1&\cellcolor{tu0!40}$-$0.1&{$+$0.0}&&\cellcolor{tu9!20}$+$0.1&\cellcolor{tu9!20}$+$0.1&$+$0.0&\cellcolor{tu0!40}$-$0.2&$+$0.0&{$+$0.0}&{$+$0.0}&&{$+$0.0}&\cellcolor{tu9!20}$+$0.1\\
    \bottomrule
  \end{tabular}
\caption{\textbf{(Average) top-1 accuracy for default FroFA} on \textit{our} ILSVRC-2012 test set. Absolute gains to the MAP baseline are reported. We use the L/16 JFT-3B base setup (\cf \cref{sec:exp_results}). In total, we investigate eighteen FroFAs, categorized into \textit{geometric}, \textit{crop \& drop}, \textit{stylistic}, and \textit{other}. We sweep across a base sweep (\cf \cref{sec:training_details}) and the respective augmentation sweep (\cf Appendix, \cref{sec:appendix_augmentation}) to first find the best setting on \textit{our} ILSVRC-2012 validation set. Each shot is sampled five times, except for {JPEG}, {sharpness}, and {solarize} (marked with `*'). We highlight \colorbox{tu0!20}{deterioration} by \colorbox{tu0!20}{shades of red} and \colorbox{tu9!20}{improvement} by \colorbox{tu9!20}{shades of green}. Best three FroFAs are boldfaced.}
  \label{tab:single_augmentation}
\end{table*}

As a next step, we investigate the effect of adding a single FroFA to the MAP baseline setting. We first focus on the default FroFA formulation which uses a single randomly sampled value per input (\cf \cref{sec:frofa}). Results are shown in \cref{tab:single_augmentation} where we report gains with respect to the MAP baseline using eighteen distinct FroFAs categorized into geometric, crop \& drop, stylistic, and other.

\textbf{Geometric}: Interestingly, all geometric augmentations consistently lead to worse performance across all settings.

\begin{figure*}
  \centering
  \begin{subfigure}{\columnwidth}
%trim={left bottom right top}
\includegraphics[width=\columnwidth]{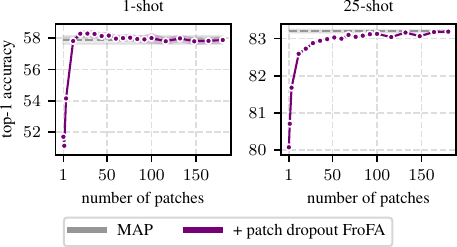}
    \caption{Patch dropout FroFA}
    \label{fig:patch_dropout}
  \end{subfigure}
  \hfill
  \begin{subfigure}{\columnwidth}
    \includegraphics[width=\columnwidth]{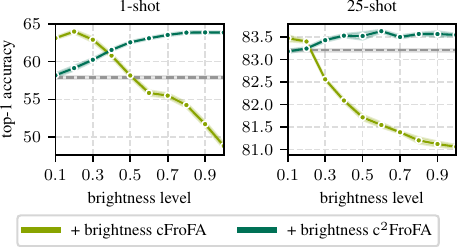}
    \caption{Channel variants (c/c$^2$) of brightness FroFA}
    \label{fig:brightness-comparison}
  \end{subfigure}
  \caption{\textbf{Average top-1 accuracy for FroFA variants} on \textit{our} ILSVRC-2012 test set. We use the L/16 JFT-3B base setup (\cf \cref{sec:exp_results}). We sweep across a base sweep (\cf \cref{sec:training_details}) to first find the best setting on \textit{our} ILSVRC-2012 validation set for each FroFA operation point (\cf Appendix, \cref{sec:appendix_augmentation}). Shaded areas indicate standard errors collected via sampling each shot five times.}
  \label{fig:brightness+pd-comparison}
\end{figure*}

\textbf{Crop \& drop}: 
A simple crop or a resized crop yield a significant performance boost in the 1-shot setting of +3.0\% and +1.9\% absolute, respectively. Further, patch dropout provides modest gains in the 1-shot regime. Dropping patches is related to training efficiency, so we investigate this further. \cref{fig:patch_dropout} shows the top-1 accuracy on 1- and 25-shot as a function of number of patches. More results can be found in Appendix, \cref{sec:appendix_exp_patch_drop_and_brightness}. Similar to observations by Liu \etal \cite{Liu2023} we can randomly drop a large fraction of patches ($>$50\%) without loosing performance. A key difference is that Liu \etal only investigated the effect in the image space, while we provide evidence that patch dropout also transfers to the feature space. Finally, inception crop does not improve performance.

\textbf{Stylistic}: The largest gains can be observed when employing a stylistic FroFA, in particular brightness, contrast, and posterize. We identified brightness as the best performing FroFA with absolute gains of 4.8\% on 1-shot, 1.1\% on 5-shot, and up to 0.6\% on 10-shot.

\textbf{Other}: Neither JPEG nor mixup yield performance gains but rather more or less worsen the performance.

\subsection{Channel FroFA}

\begin{table}
\setlength{\tabcolsep}{2.8pt}
\footnotesize
  \centering
  \begin{tabular}{@{}lccccccccccc@{}}
    \toprule
    &&&\multicolumn{3}{c}{Brightness}&&\multicolumn{2}{c}{Contrast}&&\multicolumn{2}{c}{Posterize}\\
    Shots & MAP && \rotatebox{0}{{\_}}&\rotatebox{0}{{c}}&\rotatebox{0}{{c$^2$}}&&\rotatebox{0}{{\_}}&\rotatebox{0}{{c}}&&\rotatebox{0}{{\_}}&\rotatebox{0}{{c}}\\
    \midrule
    1 &  57.9&& $+$4.8&\textbf{$+$5.9}&\textbf{$+$6.1}&&\textbf{$+$2.8}&$+$2.5&&$+$3.7&\textbf{$+$5.9}\\
    5 &  78.8 && $+$1.1&\textbf{$+$1.5}&\textbf{$+$1.6}&&\textbf{$+$0.8}&$+$0.0&&\textbf{$+$0.8}&\textbf{$+$0.8}\\
    10 &  80.9 && $+$0.6&\textbf{$+$1.1}&\textbf{$+$0.9}&&\textbf{$+$0.6}&$+$0.0&&\textbf{$+$0.6}&\textbf{$+$0.5}\\
    25 &  83.2 && $+$0.1&\textbf{$+$0.4}&\textbf{$+$0.3}&&\textbf{$+$0.1}&\textbf{$-$0.1}&&\textbf{$+$0.0}&\textbf{$+$0.0}\\
    \bottomrule
  \end{tabular}
  \caption{\textbf{Average top-1 accuracy for a selection of default (\_) and channel (c/c$^2$) FroFA} on \textit{our} ILSVRC-2012 test set. Absolute gains to the MAP baseline are reported. We use the L/16 JFT-3B base setup (\cf \cref{sec:exp_results}). We sweep across a base sweep (\cf \cref{sec:training_details}) and the respective augmentation sweep (\cf Appendix, \cref{sec:appendix_augmentation}) to first find the best setting on \textit{our} ILSVRC-2012 validation set. Each shot is sampled five times. The best results per shot \textit{and} FroFA are boldfaced (multiple ones if close, \ie, $\pm$0.2).}
  \label{tab:feature-wise}
\end{table}

Next, we investigate channel FroFA (cFroFA) for brightness, contrast, and posterize. Results are shown in \cref{tab:feature-wise}, where we report absolute gains with respect to the MAP baseline. First, contrast cFroFA worsens performance across all shots. Second, posterize cFroFA improves performance on 1-shot from +3.7\% to +5.9\% while maintaining performance on all other shots. Lastly, brightness cFroFA significantly improves performance across all shots, \ie, from +4.8\% to +5.9\% on 1-shot, from +1.1\% to +1.5\% on 5-shot, from +0.6\% to +1.1\% on 10-shot, and from +0.1\% to +0.4\% on 25-shot.

Giving the strong improvements for brightness cFroFA, we further test brightness c$^2$FroFA (see \cref{tab:feature-wise}). On a first look, both variants perform equally well. In \cref{fig:brightness-comparison}, we further report the top-1 accuracy on 1-shot and 25-shot as a function of the brightness augmentation level. Results across other shots are similar and can be found in Appendix, \cref{sec:appendix_exp_patch_drop_and_brightness}. We clearly observe that brightness cFroFA is much more sensitive to the brightness level than brightness c$^2$FroFA. Aross all shots, brightness cFroFA only works well for small brightness levels (0.1 to 0.5), while the c$^2$FroFA variant performs better than the MAP baseline across the board. We attribute the better sensitivity properties of brightness c$^2$FroFA to the channel-wise mappings \eqref{eq:f_to_x_specific}, \eqref{eq:x_to_f_specific} since this is the only change between cFroFA and c$^2$FroFA. We did not a observe similar effect when switching from cFroFA posterize to c$^2$FroFA posterize.

\subsection{Sequential FroFA}

Finally, out of our best three augmentations, \ie, brightness c$^2$FroFA (B-c$^2$), contrast FroFA (C), and posterize cFroFA (P-c), we combine two of them sequentially. We end up with a total of six combinations. \cref{tab:sequential} compares the performance of these six combinations against our prior best (B-c$^2$). On 1-shot, (B-c$^2$$\rightarrow$P-c) significantly outperforms (B-c$^2$), improving absolute gains from 6.1\% to 7.7\%, while maintaining performance on other shots. We conclude that advanced FroFA protocols may further improve performance. As an initial investigation, we applied variations of RandAugment and TrivialAugment using our best three FroFAs (\cf \cref{tab:feature-wise}), however, with limited success. We include results in the Appendix, \cref{sec:appendix_exp_advanced_protocols}, and leave a deeper investigation to future works.

% \subsection{Summary}
% We conclude our first set of experiments by emphasizing some highlights. [...] \TODO{Repeat the highlights. This can be seen as a sort of pit stop section and should point the reader (a) to the most important highlights (consistent to the ones mentioned in the introduction [and abnstract]) and motivate the next section where we want to check whether the highlights (somewhat) transfer across models and datasets.}

\section{FroFA on More Datasets and Architectures}

How well does our best non-sequential augmentation strategy (brightness c$^2$FroFA) transfer across multiple dataset and architectures settings? In \cref{ssec:other_architectures,ssec:i21k_pretraining,ssec:transfer_datasets}, we report results on seven other downstream few-shot datasets, two additional architectures, and two additional pretraining setups, respectively. This time, however, we also incorporate \textit{weight decay in all MAP-based models}. Further, in \cref{ssec:other_architectures,ssec:i21k_pretraining}, we solely focus on the improvements over the MAP baseline and include a discussion on the improvements over the linear probe baseline in \cref{sec:linear_probe,ssec:transfer_datasets}.

\subsection{Transfer to Other Downstream Datasets}
\label{ssec:transfer_datasets}

In \cref{tab:additional_transfer}, we report results on seven additional transfer datasets, \ie, CIFAR10, CIFAR100, DMLab, DTD, Resisc45, SUN397, and SVHN. We compare the weight-decayed MAP and L2-regularized linear probe baseline to our approach, \ie, weight-decayed MAP combined with brightness $c^2$FroFA (MAP\textsuperscript{wd} + FroFA). We observe that across almost all shots and transfer datasets, MAP\textsuperscript{wd} + FroFA shows the best results. Moreover, MAP\textsuperscript{wd} + FroFA outperforms L2-regularized linear probe with only one exception, \ie, SUN397 (1-shot). With respect to the mean across all seven datasets, MAP\textsuperscript{wd} + FroFA is significantly better than MAP\textsuperscript{wd}, with improvements ranging from +4.4\% absolute on 1-shot to +1.0\% absolute on 25-shot.

%double digits
% \begin{table}
% \setlength{\tabcolsep}{2.8pt}
% \footnotesize
%   \centering
%   \begin{tabular}{@{}lcccccccccccc@{}}
%     \toprule
%     Shots & MAP && \rotatebox{0}{{B-c$^2$}}&&\rotatebox{90}{{B-c$^2$$\rightarrow$C}}&\rotatebox{90}{{C$\rightarrow$ B-c$^2$}}&&\rotatebox{90}{{B-c$^2$$\rightarrow$P-c}}&\rotatebox{90}{{P-c$\rightarrow$ B-c$^2$}}&&\rotatebox{90}{{C$\rightarrow$P-c}}&\rotatebox{90}{{P-c$\rightarrow$C}}\\
%     \midrule
%     1 &  57.93&&\cellcolor{gray!20}+6.05&&+3.95&+2.66&&+\cellcolor{gray!60}7.72&+5.17&&+4.97&+3.11\\
%     5 &  78.77 && \cellcolor{gray!60}+1.60&& +1.50&+0.19&&\cellcolor{gray!20}+1.54&+0.40&&+1.33&-0.02\\
%     10 &  80.91 && +0.93&&\cellcolor{gray!60}+1.17&+0.06&&\cellcolor{gray!20}+1.01&+0.12&&+0.89&+0.32\\
%     25 &  83.19 &&\cellcolor{gray!20}+0.28&&\cellcolor{gray!60}+0.38&-0.70&&+0.23&-0.45&&+0.17&-0.38\\
%     \bottomrule
%   \end{tabular}
%   \caption{\textbf{Top-1 accuracy for a sequential FroFA protocol} on \textit{our} ILSVRC-2012 test set. Absolute gains to the MAP baseline are reported. We use the L/16 JFT-3B base setup (\cf \cref{sec:exp_results}). We combine the best settings of brightness c$^2$FroFA (B-c$^2$), contrast FroFA (C), and posterize cFroFA (P-c) sequentially (two at a time, order indicated by `$\uparrow$'). We sweep across a base sweep (\cf \cref{sec:training_details}) to first find the best setting on \textit{our} ILSVRC-2012 validation set. The \colorbox{gray!60}{best} and \colorbox{gray!20}{second-best} results per shot and FroFA setting are highlighted.}
%   \label{tab:sequential}
% \end{table}
%single digit
\begin{table}
\setlength{\tabcolsep}{2.8pt}
\footnotesize
  \centering
  \begin{tabular}{@{}lcccccccccccc@{}}
    \toprule
    Shots & MAP && \rotatebox{0}{{B-c$^2$}}&&\rotatebox{90}{{B-c$^2$$\rightarrow$C}}&\rotatebox{90}{{C$\rightarrow$ B-c$^2$}}&&\rotatebox{90}{{B-c$^2$$\rightarrow$P-c}}&\rotatebox{90}{{P-c$\rightarrow$ B-c$^2$}}&&\rotatebox{90}{{C$\rightarrow$P-c}}&\rotatebox{90}{{P-c$\rightarrow$C}}\\
    \midrule
    1 &  57.9&&$+$6.1&&$+$4.0&$+$2.7&&$+$\textbf{7.7}&$+$5.2&&$+$5.0&$+$3.1\\
    5 &  78.8 && \textbf{$+$1.6}&& \textbf{$+$1.5}&$+$0.2&&\textbf{$+$1.5}&$+$0.4&&$+$1.3&$+$0.0\\
    10 &  80.9 && $+$0.9&&\textbf{$+$1.2}&$+$0.1&&\textbf{$+$1.0}&$+$0.1&&$+$0.9&$+$0.3\\
    25 &  83.2 &&\textbf{$+$0.3}&&\textbf{$+$0.4}&$-$0.7&&\textbf{$+$0.2}&$-$0.5&&$+$0.2&$-$0.4\\
    \bottomrule
  \end{tabular}
  \caption{\textbf{Average top-1 accuracy for a sequential FroFA protocol} on \textit{our} ILSVRC-2012 test set. Absolute gains to the MAP baseline are reported. We use the L/16 JFT-3B base setup (\cf \cref{sec:exp_results}). We combine the best settings of brightness c$^2$FroFA (B-c$^2$), contrast FroFA (C), and posterize cFroFA (P-c) sequentially (two at a time, order indicated by `$\uparrow$'). We sweep across a base sweep (\cf \cref{sec:training_details}) to first find the best setting on \textit{our} ILSVRC-2012 validation set. Each shot is sampled five times. The best results per shot are boldfaced (multiple ones if close, \ie, $\pm$0.2).}
  \label{tab:sequential}
\end{table}

\begin{table}
\setlength{\tabcolsep}{2.8pt}
\footnotesize
\renewcommand{\arraystretch}{0.9}
  \centering
  \begin{tabular}{@{}llcccc@{}}
    \toprule
    Trans.\ dataset&Method & 1-shot & 5-shot & 10-shot& 25-shot \\
    \midrule
    \multirow{3}{*}{\rotatebox{0}{CIFAR10}}&MAP\textsuperscript{wd} & \underline{85.1} & \underline{96.7} & 97.1 & \underline{97.5} \\
    &Linear probe& 80.9 & 94.1 & \underline{96.7}  & \underline{97.3}  \\
    &MAP\textsuperscript{wd} + FroFA & \textbf{93.8} & \textbf{97.6} & \textbf{97.8}  & \textbf{97.8} \\
    \midrule
     \multirow{3}{*}{\rotatebox{0}{CIFAR100}}&MAP\textsuperscript{wd} & \underline{63.1}& \underline{82.7}&\underline{85.5}&\underline{86.8}\\
    &Linear probe & 58.4 & 80.9 & 83.8& 85.1\\
    &MAP\textsuperscript{wd} + FroFA & \textbf{67.8} & \textbf{84.0} & \textbf{86.2} & \textbf{87.1}\\
    \midrule
    \multirow{3}{*}{\rotatebox{0}{DMLab}}&MAP\textsuperscript{wd} & \underline{24.4}& \textbf{30.3}& \textbf{30.2} & \underline{36.5} \\
    &Linear probe &  24.0 & 26.3 &\underline{25.6} & 30.9 \\
    &MAP\textsuperscript{wd} + FroFA & \textbf{27.1} & \underline{29.4} & \textbf{30.3} & \textbf{36.8} \\
    \midrule
    \multirow{3}{*}{\rotatebox{0}{DTD}}&MAP\textsuperscript{wd} &  \underline{49.2} & \underline{68.2} & \underline{74.1}& \underline{80.8}\\
    &Linear probe & 46.9 & 65.9& 71.3 & 77.3 \\
    &MAP\textsuperscript{wd} + FroFA & \textbf{53.5} & \textbf{70.7} & \textbf{76.1} & \textbf{82.2}\\
    \midrule
    \multirow{3}{*}{\rotatebox{0}{Resisc45}}&MAP\textsuperscript{wd} & 63.2& 86.9&\textbf{89.8}&90.7\\
    &Linear probe & \underline{67.1} & \underline{85.6}& \underline{88.2}& \underline{91.0}\\
    &MAP\textsuperscript{wd} + FroFA & \textbf{67.6}& \textbf{87.2}&\textbf{89.7}&\textbf{91.5}\\
    \midrule
    \multirow{3}{*}{\rotatebox{0}{SUN397}}&MAP\textsuperscript{wd} &  51.3 & \underline{73.5} & 77.7 &\underline{80.3}\\
    &Linear probe &  \textbf{56.7}& 70.9 & \underline{75.6}& 78.6\\
    &MAP\textsuperscript{wd} + FroFA &  \underline{56.2} & \textbf{75.9} &\textbf{78.9}&\textbf{81.2}\\
    \midrule
    \multirow{3}{*}{\rotatebox{0}{SVHN}}&MAP\textsuperscript{wd} & \underline{20.7} & \underline{23.9} & \underline{30.2} & \underline{47.4} \\
    &Linear probe & 11.8 & 15.0 & 18.7 & 21.5\\
    &MAP\textsuperscript{wd} + FroFA & \textbf{21.8} & \textbf{31.0} & \textbf{43.5} & \textbf{50.3} \\
    \midrule
    \midrule
    \multirow{3}{*}{\rotatebox{0}{Mean}}&MAP\textsuperscript{wd} & \underline{51.0} & \underline{66.0} & \underline{69.2}  & \underline{74.3} \\
    &Linear probe &  49.1 &  62.7 & 65.7  & 68.8 \\
    &MAP\textsuperscript{wd} + FroFA & \textbf{55.4} & \textbf{68.0} & \textbf{71.8} & \textbf{75.3} \\
    \bottomrule
  \end{tabular}
  \caption{\textbf{Top-1 accuracy of our best FroFA for additional transfer datasets} using a JFT-3B L/16 model. Results are reported on the respective test set (\cf \cref{sec:appendix_datasets}). We compare results to a weight-decayed MAP baseline, \ie, MAP\textsuperscript{wd}, and an L2-regularized linear probe. Depending on the setting, we sweep across a base, \cf \cref{sec:training_details}, a weight decay or L2 decay, \cf \cref{sec:baselines}, and a brightness level sweep, \cf \cref{sec:appendix_augmentation}, to first find the best setting on the respective validation set. Per shot and dataset, the best result is boldfaced while the second-best result is underlined (multiple ones if close, \ie, $\pm$0.2).}
  \label{tab:additional_transfer}
\end{table}
\begin{figure*}
  \centering
  \begin{subfigure}{\textwidth}
\includegraphics{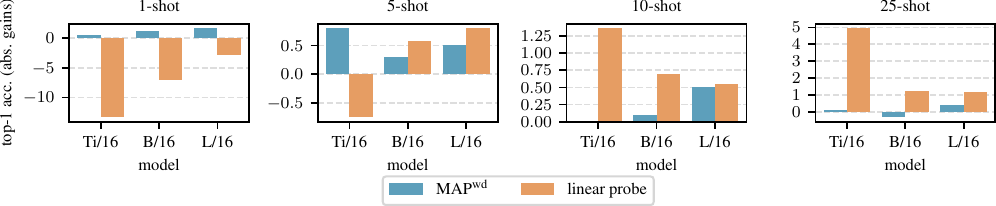}
    \caption{JFT-3B}
    \label{fig:jft3b_all}
  \end{subfigure}
  \hfill
  \begin{subfigure}{\textwidth}
\includegraphics{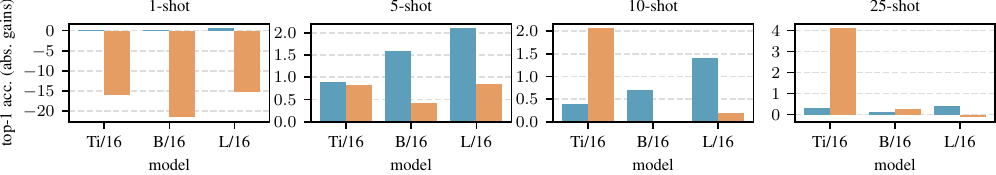}
    \caption{ImageNet21k}
    \label{fig:i21k_all}
  \end{subfigure}
  \caption{\textbf{Average top-1 accuracy of brightness c$^2$FroFA for JFT-3B \subref{fig:jft3b_all} and ImageNet-21k \subref{fig:i21k_all} models} on \textit{our} ILSVRC-2012 test set trained on few-shotted ILSVRC-2012 training sets. Absolute gains to the weight-decayed MAP, \ie MAP\textsuperscript{wd}, and L2-regularized linear probe baseline are reported. Depending on the setting, we sweep across a base, \cf \cref{sec:training_details}, a weight decay or L2 decay, \cf \cref{sec:baselines}, and a brightness level sweep, \cf \cref{sec:appendix_augmentation}, to first find the best setting on \textit{our} ILSVRC-2012 validation set for each model.}
  \label{fig:jft3b_and_i21k_all}
\end{figure*}

\cref{fig:first_page}, left, displays the absolute accuracy gains averaged across all eight transfer datasets, including ILSVRC-2012. As before, our approach, \ie, MAP\textsuperscript{wd} + FroFA, yields the best results across all shots. We further observe that the gains decrease with higher shots which aligns with our previous observations.

\subsection{Transfer to Other Architectures}
\label{ssec:other_architectures}
We employ brightness c$^2$FroFA on two other JFT-pretrained models, namely Ti/16 and B/16. In \cref{fig:jft3b_all}, we report improvements in top-1 accuracy with respect to the weight-decayed MAP baseline. Across all shots and model architectures, incorporating FroFA either \textit{maintains or improves performance}, except for B/16, 25-shot. Given that larger models tend to be more prone to overfitting in the 1-shot setting, we observe increasing improvements from FroFA when scaling the architecture. With a higher number of shots, the observed improvements over the baseline model become smaller. We attribute this to the strong baseline performance leaving lesser headroom for improvements.
We refer to the Appendix, \cref{sec:appendix_exp_frofa_transfer}, for the exact values.

\subsection{Transfer to Other Pretraining Setups}
\label{ssec:i21k_pretraining}
\textbf{ImageNet-21k}: In \cref{fig:i21k_all}, we report improvements in top-1 accuracy with respect to the weight-decayed MAP baseline for ImageNet-21k-pretrained Ti/16, B/16, and L/16. Consistent with our JFT-3B observations, across all shots and model architectures, incorporating FroFA either \textit{maintains or improves performance}. The improvements diminish as the number of shots increases. This trend is likely due to the higher baseline accuracies at higher shot counts.
We again refer to the Appendix, \cref{sec:appendix_exp_frofa_transfer}, for the exact values.

\textbf{WebLI and SigLIP}:
We also tested an L/16 model with sigmoid language-image pretraining (SigLIP), following \cite{Zhai2023}. We report the absolute accuracy gains averaged across eight datasets. The results are shown in \cref{fig:first_page}, right. From the results we can conclude that our FroFA setting also transfers to language-image pretrained models further emphasizing its generalizability.

\subsection{Linear Probe Comparison on ILSVRC-2012}\label{sec:linear_probe}
\label{ssec:linear_probe_comparison}

 We will now look at \cref{fig:jft3b_all,fig:i21k_all}, but discuss gains with respect to the L2-regularized linear probe baseline. We start with models pretrained on JFT-3B (\cf \cref{fig:jft3b_all}). On 1-shot, we observe that we lack behind linear probe but can close the gap by scaling up the model size. On 5- to 25-shot, with the exception of Ti/16 on 5-shot, brightness c$^2$FroFA significantly outperforms the linear probe baseline.

On ImageNet-21k (\cf \cref{fig:i21k_all}), we observe even larger gaps to linear probe on 1-shot (up to -20\% absolute). However, similar to results on JFT-3B, performance on 5- to 25-shot improves significantly over linear probe or at worst stays the same.
\section{Conclusions} \label{sec:conclusions}
We investigated eighteen frozen feature augmentations (FroFAs) along three axes: model size, pretraining and transfer few-shot dataset. We show that a training with FroFAs, in particular stylistic ones, gives large improvements upon a representative baseline across all shots. In addition, per-channel variants further improve performance, e.g., by 1.6\% absolute in the ILSVRC-2012 5-shot setting. Finally, we were able to show that our results transfer. Averaged results across seven downstream tasks show that using a variant of brightness FroFA improves by 4.4\% absolute upon the same representative baseline in the 1-shot setting.
{
    \small
    \bibliographystyle{ieeenat_fullname}
    \bibliography{main}

\begin{thebibliography}{59}
\providecommand{\natexlab}[1]{#1}
\providecommand{\url}[1]{\texttt{#1}}
\expandafter\ifx\csname urlstyle\endcsname\relax
  \providecommand{\doi}[1]{doi: #1}\else
  \providecommand{\doi}{doi: \begingroup \urlstyle{rm}\Url}\fi

\bibitem[{A. Krizhevsky}(2009)]{Krizhevsky2009}
{A. Krizhevsky}.
\newblock {Learning Multiple Layers of Features from Tiny Images}, 2009.

\bibitem[Beattie et~al.(2016)Beattie, Leibo, Teplyashin, Ward, Wainwright, K{\"{u}}ttler, Lefrancq, Green, Vald{\'{e}}s, Sadik, Schrittwieser, Anderson, York, Cant, Cain, Bolton, Gaffney, King, Hassabis, Legg, and Petersen]{Beattie2016}
Charles Beattie, Joel~Z. Leibo, Denis Teplyashin, Tom Ward, Marcus Wainwright, Heinrich K{\"{u}}ttler, Andrew Lefrancq, Simon Green, V{\'{\i}}ctor Vald{\'{e}}s, Amir Sadik, Julian Schrittwieser, Keith Anderson, Sarah York, Max Cant, Adam Cain, Adrian Bolton, Stephen Gaffney, Helen King, Demis Hassabis, Shane Legg, and Stig Petersen.
\newblock {DeepMind Lab}.
\newblock \emph{arXiv}, 1612.03801:\penalty0 1--11, 2016.

\bibitem[Beyer et~al.(2022)Beyer, Zhai, and Kolesnikov]{Beyer2022}
Lucas Beyer, Xiaohua Zhai, and Alexander Kolesnikov.
\newblock {Better Plain ViT Baselines for ImageNet-1k}.
\newblock \emph{arXiv}, 2205.01580:\penalty0 1--3, 2022.

\bibitem[Chen et~al.(2022)Chen, Ge, Tong, Wang, Song, Wang, and Luo]{Chen2022}
Shoufa Chen, Chongjian Ge, Zhan Tong, Jiangliu Wang, Yibing Song, Jue Wang, and Ping Luo.
\newblock {AdaptFormer: Adapting Vision Transformers for Scalable Visual Recognition}.
\newblock In \emph{Proc.\ of NeurIPS}, pages 16664--16678, New Orleans, LA, USA, 2022.

\bibitem[Chen et~al.(2023)Chen, Wang, Changpinyo, Piergiovanni, Padlewski, Salz, Goodman, Grycner, Mustafa, Beyer, Kolesnikov, Puigcerver, Ding, Rong, Akbari, Mishra, Xue, Thapliyal, Bradbury, Kuo, Seyedhosseini, Jia, Ayan, Riquelme, Steiner, Angelova, Zhai, Houlsby, and Soricut]{Chen2023}
Xi Chen, Xiao Wang, Soravit Changpinyo, AJ Piergiovanni, Piotr Padlewski, Daniel Salz, Sebastian Goodman, Adam Grycner, Basil Mustafa, Lucas Beyer, Alexander Kolesnikov, Joan Puigcerver, Nan Ding, Keran Rong, Hassan Akbari, Gaurav Mishra, Linting Xue, Ashish Thapliyal, James Bradbury, Weicheng Kuo, Mojtaba Seyedhosseini, Chao Jia, Burcu~Karagol Ayan, Carlos Riquelme, Andreas Steiner, Anelia Angelova, Xiaohua Zhai, Neil Houlsby, and Radu Soricut.
\newblock {PaLI: A Jointly-Scaled Multilingual Language-Image Model}.
\newblock In \emph{Proc.\ of ICLR}, pages 1--33, Kigali, Rwanda, 2023.

\bibitem[Cheng et~al.(2017)Cheng, Han, and Lu]{Cheng2017}
Gong Cheng, Junwei Han, and Xiaoqiang Lu.
\newblock {Remote Sensing Image Scene Classification: Benchmark and State of the Art}.
\newblock \emph{Proc.\ IEEE}, 105\penalty0 (10):\penalty0 1865--1883, 2017.

\bibitem[Chollet(2017)]{Chollet2017}
Fran\c{c}ois Chollet.
\newblock {Xception: Deep Learning With Depthwise Separable Convolutions}.
\newblock In \emph{Proc.\ of CVPR}, pages 1063--6919, Honolulu, HI, USA, 2017.

\bibitem[Cimpoi et~al.(2014)Cimpoi, Maji, Kokkinos, Mohamed, and Vedaldi]{Cimpoi2014}
Mircea Cimpoi, Subhransu Maji, Iasonas Kokkinos, Sammy Mohamed, and Andrea Vedaldi.
\newblock {Describing Textures in the Wild}.
\newblock In \emph{Proc.\ of CVPR}, pages 3606--3613, Columbus, OH, USA, 2014.

\bibitem[Cubuk et~al.(2019)Cubuk, Zoph, Mane, Vasudevan, and Le]{Cubuk2019}
Ekin~D. Cubuk, Barret Zoph, Dandelion Mane, Vijay Vasudevan, and Quoc~V. Le.
\newblock {AutoAugment: Learning Augmentation Strategies From Data}.
\newblock In \emph{Proc.\ of CVPR}, pages 113--123, Long Beach, CA, USA, 2019.

\bibitem[Cubuk et~al.(2020)Cubuk, Zoph, Shlens, and Le]{Cubuk2020}
Ekin~Dogus Cubuk, Barret Zoph, Jon Shlens, and Quoc Le.
\newblock {RandAugment: Practical Automated Data Augmentation with a Reduced Search Space}.
\newblock In \emph{Proc.\ of NeurIPS}, pages 18613--18624, virtual, 2020.

\bibitem[Dai et~al.(2021)Dai, Liu, Le, and Tan]{Dai2021}
Zihang Dai, Hanxiao Liu, Quoc~V. Le, and Mingxing Tan.
\newblock {CoAtNet: Marrying Convolution and Attention for All Data Sizes}.
\newblock In \emph{Proc.\ of NeurIPS}, pages 3965--3977, virtual, 2021.

\bibitem[Dehghani et~al.(2022)Dehghani, Gritsenko, Arnab, Minderer, and Tay]{Dehghani2022}
Mostafa Dehghani, Alexey Gritsenko, Anurag Arnab, Matthias Minderer, and Yi Tay.
\newblock {Scenic: A JAX Library for Computer Vision Research and Beyond}.
\newblock In \emph{Proc.\ of CVPR}, pages 21393--21398, New Orleans, LA, USA, 2022.

\bibitem[Dehghani et~al.(2023)Dehghani, Djolonga, Mustafa, Padlewski, Heek, Gilmer, Steiner, Caron, Geirhos, Alabdulmohsin, Jenatton, Beyer, Tschannen, Arnab, Wang, Riquelme~Ruiz, Minderer, Puigcerver, Evci, Kumar, Steenkiste, Elsayed, Mahendran, Yu, Oliver, Huot, Bastings, Collier, Gritsenko, Birodkar, Vasconcelos, Tay, Mensink, Kolesnikov, Pavetic, Tran, Kipf, Lucic, Zhai, Keysers, Harmsen, and Houlsby]{Dehghani2023}
Mostafa Dehghani, Josip Djolonga, Basil Mustafa, Piotr Padlewski, Jonathan Heek, Justin Gilmer, Andreas~Peter Steiner, Mathilde Caron, Robert Geirhos, Ibrahim Alabdulmohsin, Rodolphe Jenatton, Lucas Beyer, Michael Tschannen, Anurag Arnab, Xiao Wang, Carlos Riquelme~Ruiz, Matthias Minderer, Joan Puigcerver, Utku Evci, Manoj Kumar, Sjoerd~Van Steenkiste, Gamaleldin~Fathy Elsayed, Aravindh Mahendran, Fisher Yu, Avital Oliver, Fantine Huot, Jasmijn Bastings, Mark Collier, Alexey~A. Gritsenko, Vighnesh Birodkar, Cristina~Nader Vasconcelos, Yi Tay, Thomas Mensink, Alexander Kolesnikov, Filip Pavetic, Dustin Tran, Thomas Kipf, Mario Lucic, Xiaohua Zhai, Daniel Keysers, Jeremiah~J. Harmsen, and Neil Houlsby.
\newblock {Scaling Vision Transformers to 22 Billion Parameters}.
\newblock In \emph{Proc.\ of ICML}, pages 7480--7512, Honolulu, HI, USA, 2023.

\bibitem[Deng et~al.(2009)Deng, Dong, Socher, Li, Li, and Fei-Fei]{Deng2009}
Jia Deng, Wei Dong, Richard Socher, Li-Jia Li, Kai Li, and Li Fei-Fei.
\newblock {{ImageNet}: {A} Large-Scale Hierarchical Image Database}.
\newblock In \emph{Proc.\ of CVPR}, pages 248--255, Miami, FL, USA, 2009.

\bibitem[DeVries and Taylor(2017)]{DeVries2017}
Terrance DeVries and Graham~W. Taylor.
\newblock {Dataset Augmentation in Feature Space}.
\newblock In \emph{Proc.\ of ICLR - Workshops}, pages 1--12, Toulon, France, 2017.

\bibitem[Dosovitskiy et~al.(2021)Dosovitskiy, Beyer, Kolesnikov, Weissenborn, Zhai, Unterthiner, Dehghani, Minderer, Heigold, Gelly, Uszkoreit, and Houlsby]{Dosovitskiy2021}
Alexey Dosovitskiy, Lucas Beyer, Alexander Kolesnikov, Dirk Weissenborn, Xiaohua Zhai, Thomas Unterthiner, Mostafa Dehghani, Matthias Minderer, Georg Heigold, Sylvain Gelly, Jakob Uszkoreit, and Neil Houlsby.
\newblock {An Image is Worth 16x16 Words: Transformers for Image Recognition at Scale}.
\newblock In \emph{Proc.\ of ICLR}, pages 1--21, virtual, 2021.

\bibitem[Gao et~al.(2023)Gao, Geng, Zhang, Ma, Fang, Zhang, Li, and Qiao]{Gao2023}
Peng Gao, Shijie Geng, Renrui Zhang, Teli Ma, Rongyao Fang, Yongfeng Zhang, Hongsheng Li, and Yu Qiao.
\newblock {CLIP-Adapter: Better Vision-Language Models with Feature Adapters}.
\newblock \emph{Int. J. Comput. Vis.}, pages 1--15, 2023.

\bibitem[Grave et~al.(2017)Grave, Joulin, and Usunier]{Grave2017}
Edouard Grave, Armand Joulin, and Nicolas Usunier.
\newblock {Improving Neural Language Models with a Continuous Cache}.
\newblock In \emph{Proc.\ of ICLR}, pages 1--9, Toulon, France, 2017.

\bibitem[He et~al.(2023)He, Li, Zhang, Yang, and Wang]{He2023}
Xuehai He, Chuanyuan Li, Pengchuan Zhang, Jianwei Yang, and Xin~Eric Wang.
\newblock {Parameter-Efficient Model Adaptation for Vision Transformers}.
\newblock In \emph{Proc.\ of AAAI}, pages 817--825, Washington, DC, USA, 2023.

\bibitem[Hendrycks et~al.(2020)Hendrycks, Mu, Cubuk, Zoph, Gilmer, and Lakshminarayanan]{Hendrycks2020}
Dan Hendrycks, Norman Mu, Ekin~D. Cubuk, Barret Zoph, Justin Gilmer, and Balaji Lakshminarayanan.
\newblock {AugMix: A Simple Data Processing Method to Improve Robustness and Uncertainty}.
\newblock In \emph{Proc.\ of ICLR}, pages 1--15, Virtual, 2020.

\bibitem[Hinton et~al.(2014)Hinton, Vinyals, and Dean]{Hinton2014}
Geoffrey Hinton, Oriol Vinyals, and Jeff Dean.
\newblock {Distilling Knowledge in a Neural Network}.
\newblock In \emph{proc.\ of NIPS - Workshops}, pages 1--9, Montr{\' e}al, QC, Canada, 2014.
\newblock (`NIPS' was renamed to `NeurIPS' after 2018).

\bibitem[Houlsby et~al.(2019)Houlsby, Giurgiu, Jastrzebski, Morrone, de~Laroussilhe, Gesmundo, Attariyan, and Gelly]{Houlsby2019}
Neil Houlsby, Andrei Giurgiu, Stanislaw Jastrzebski, Bruna Morrone, Quentin de Laroussilhe, Andrea Gesmundo, Mona Attariyan, and Sylvain Gelly.
\newblock {Parameter-Efficient Transfer Learning for NLP}.
\newblock In \emph{Proc.\ of ICML}, pages 2790--2799, Long Beach, CA, USA, 2019.

\bibitem[Hu et~al.(2022)Hu, Shen, Wallis, Allen-Zhu, Li, Wang, Wang, and Chen]{Hu2022}
Edward~J. Hu, Yelong Shen, Phillip Wallis, Zeyuan Allen-Zhu, Yuanzhi Li, Shean Wang, Lu Wang, and Weizhu Chen.
\newblock {LoRA: Low-Rank Adaptation of Large Language Models}.
\newblock In \emph{Proc.\ of ICLR}, pages 1--13, virtual, 2022.

\bibitem[Jia et~al.(2022)Jia, Tang, Chen, Cardie, Belongie, Hariharan, and Lim]{Jia2022}
Menglin Jia, Luming Tang, Bor-Chun Chen, Claire Cardie, Serge Belongie, Bharath Hariharan, and Ser-Nam Lim.
\newblock {Visual Prompt Tuning}.
\newblock In \emph{Proc.\ of ECCV}, pages 709--727, Tel Aviv, Israel, 2022.

\bibitem[Kingma and Ba(2015)]{Kingma2015}
Diederik~P. Kingma and Jimmy Ba.
\newblock {Adam: A Method for Stochastic Optimization}.
\newblock In \emph{Proc.\ of ICLR}, pages 1--15, San Diego, CA, USA, 2015.

\bibitem[Kolesnikov et~al.(2020)Kolesnikov, Beyer, Zhai, Puigcerver, Yung, Gelly, and Houlsby]{Kolesnikov2020}
Alexander Kolesnikov, Lucas Beyer, Xiaohua Zhai, Joan Puigcerver, Jessica Yung, Sylvain Gelly, and Neil Houlsby.
\newblock {Big Transfer (BiT): General Visual Representation Learning}.
\newblock In \emph{Proc.\ of ECCV}, pages 491--507, virtual, 2020.

\bibitem[Kossen et~al.(2023)Kossen, Collier, Mustafa, Wang, Zhai, Beyer, Steiner, Berent, Jenatton, and Kokiopoulou]{Kossen2023}
Jannik Kossen, Mark Collier, Basil Mustafa, Xiao Wang, Xiaohua Zhai, Lucas Beyer, Andreas Steiner, Jesse Berent, Rodolphe Jenatton, and Efi Kokiopoulou.
\newblock {Three Towers: Flexible Contrastive Learning with Pretrained Image Models}.
\newblock \emph{arXiv}, 2112.13492:\penalty0 1--32, 2023.

\bibitem[Kumar et~al.(2019)Kumar, Glaude, de~Lichy, and Campbell]{Kumar2019a}
Varun Kumar, Hadrien Glaude, Cyprien de Lichy, and Wlliam Campbell.
\newblock {A Closer Look At Feature Space Data Augmentation For Few-Shot Intent Classification}.
\newblock In \emph{Proc.\ of EMNLP - Workshops}, pages 1--10, Hong Kong, China, 2019.

\bibitem[Lake et~al.(2019)Lake, Salakhutdinov, and Tenenbaum]{Lake2019}
Brenden~M. Lake, Ruslan Salakhutdinov, and Joshua~B. Tenenbaum.
\newblock {The Omniglot Challenge: a 3-year Progress Report}.
\newblock \emph{Curr. Opin. Behav. Sci.}, 29:\penalty0 97--104, 2019.

\bibitem[Lee et~al.(2019)Lee, Lee, Kim, Kosiorek, Choi, and Teh]{Lee2019f}
Juho Lee, Yoonho Lee, Jungtaek Kim, Adam Kosiorek, Seungjin Choi, and Yee~Whye Teh.
\newblock {Set Transformer: A Framework for Attention-based Permutation-Invariant Neural Networks}.
\newblock In \emph{Proc.\ of ICML}, pages 3744--3753, Long Beach, CA, USA, 2019.

\bibitem[Lee et~al.(2021)Lee, Lee, and Song]{lee2021vision}
Seung~Hoon Lee, Seunghyun Lee, and Byung~Cheol Song.
\newblock {Vision Transformer for Small-size Datasets}.
\newblock \emph{arXiv}, 2112.13492:\penalty0 1--11, 2021.

\bibitem[Lester et~al.(2021)Lester, Al-Rfou, and Constant]{Lester2021}
Brian Lester, Rami Al-Rfou, and Noah Constant.
\newblock {The Power of Scale for Parameter-Efficient Prompt Tuning}.
\newblock In \emph{Proc.\ of EMNLP}, pages 3045--3059, virtual, 2021.

\bibitem[Liu et~al.(2018)Liu, Zou, Kong, Diao, Yan, Wang, Li, Jia, and You]{Liu2018a}
Xiaofeng Liu, Yang Zou, Lingsheng Kong, Zhihui Diao, Junliang Yan, Jun Wang, Site Li, Ping Jia, and Jane You.
\newblock {Data Augmentation via Latent Space Interpolation for Image Classification}.
\newblock In \emph{Proc.\ of ICPR}, pages 728--733, Beijing, China, 2018.

\bibitem[Liu et~al.(2021{\natexlab{a}})Liu, Sangineto, Bi, Sebe, Lepri, and Nadai]{liu2021efficient}
Yahui Liu, Enver Sangineto, Wei Bi, Nicu Sebe, Bruno Lepri, and Marco~De Nadai.
\newblock {Efficient Training of Visual Transformers With Small Datasets}.
\newblock In \emph{Proc.\ of NeurIPS}, pages 1--13, virtual, 2021{\natexlab{a}}.

\bibitem[Liu et~al.(2023{\natexlab{a}})Liu, Matsoukas, Strand, Azizpour, and Smith]{Liu2023}
Yue Liu, Christos Matsoukas, Fredrik Strand, Hossein Azizpour, and Kevin Smith.
\newblock {PatchDropout: Economizing Vision Transformers Using Patch Dropout}.
\newblock In \emph{Proc.\ of WACV}, pages 3942--3951, Waikoloa, HI, USA, 2023{\natexlab{a}}.

\bibitem[Liu et~al.(2021{\natexlab{b}})Liu, Lin, Cao, Hu, Wei, Zhang, Lin, and Guo]{liu2021swin}
Ze Liu, Yutong Lin, Yue Cao, Han Hu, Yixuan Wei, Zheng Zhang, Stephen Lin, and Baining Guo.
\newblock {Swin Transformer: Hierarchical Vision Transformer Using Shifted Windows}.
\newblock In \emph{Proc.\ of ICCV}, pages 10012--10022, virtual, 2021{\natexlab{b}}.

\bibitem[Liu et~al.(2023{\natexlab{b}})Liu, Tang, Shi, Zhang, Li, Shrivastava, and Wilson]{Liu2023a}
Zichang Liu, Zhiqiang Tang, Xingjian Shi, Aston Zhang, Mu Li, Anshumali Shrivastava, and Andrew~Gordon Wilson.
\newblock {Learning Multimodal Data Augmentation in Feature Space}.
\newblock In \emph{Proc.\ of ICLR}, pages 1--15, Kigali, Rwanda, 2023{\natexlab{b}}.

\bibitem[Loshchilov and Hutter(2019)]{Loshchilov2019}
Ilya Loshchilov and Frank Hutter.
\newblock {Decoupled Weight Decay Regularization}.
\newblock In \emph{Proc.\ of ICLR}, pages 1--18, New Orleans, LA, USA, 2019.

\bibitem[M\"uller and Hutter(2021)]{Mueller2021}
Samuel~G. M\"uller and Frank Hutter.
\newblock {TrivialAugment: Tuning-Free Yet State-of-the-Art Data Augmentation}.
\newblock In \emph{Proc.\ of ICCV}, pages 774--782, virtual, 2021.

\bibitem[Netzer et~al.(2011)Netzer, Wang, Coates, Bissacco, Wu, and Ng]{Netzer2011}
Yuval Netzer, Tao Wang, Adam Coates, Alessandro Bissacco, Bo Wu, and Andrew~Y. Ng.
\newblock {Reading Digits in Natural Images with Unsupervised Feature Learning}.
\newblock In \emph{proc.\ of NIPS - Workshops}, pages 1--9, Granada, Spain, 2011.
\newblock (`NIPS' was renamed to `NeurIPS' after 2018).

\bibitem[Oquab et~al.(2023)Oquab, Darcet, Moutakanni, Vo, Szafraniec, Khalidov, Fernandez, Haziza, Massa, El-Nouby, Assran, Ballas, Galuba, Howes, Huang, Li, Misra, Rabbat, Sharma, Synnaeve, Xu, Jegou, Mairal, Labatut, Joulin, and Bojanowski]{oquab2023dinov2}
Maxime Oquab, Timothée Darcet, Théo Moutakanni, Huy Vo, Marc Szafraniec, Vasil Khalidov, Pierre Fernandez, Daniel Haziza, Francisco Massa, Alaaeldin El-Nouby, Mahmoud Assran, Nicolas Ballas, Wojciech Galuba, Russell Howes, Po-Yao Huang, Shang-Wen Li, Ishan Misra, Michael Rabbat, Vasu Sharma, Gabriel Synnaeve, Hu Xu, Hervé Jegou, Julien Mairal, Patrick Labatut, Armand Joulin, and Piotr Bojanowski.
\newblock {Dinov2: Learning Robust Visual Features Without Supervision}.
\newblock \emph{arXiv}, 2304.07193:\penalty0 1--31, 2023.

\bibitem[Orhan(2018)]{Orhan2018}
Emin Orhan.
\newblock {A Simple Cache Model for Image Recognition}.
\newblock In \emph{Proc.\ of NeurIPS}, pages 10128--10137, Montréal, Canada, 2018.

\bibitem[Radford et~al.(2021)Radford, Kim, Hallacy, Ramesh, Goh, Agarwal, Sastry, Askell, Mishkin, Clark, Krueger, and Sutskever]{Radford2021}
Alec Radford, Jong~Wook Kim, Chris Hallacy, Aditya Ramesh, Gabriel Goh, Sandhini Agarwal, Girish Sastry, Amanda Askell, Pamela Mishkin, Jack Clark, Gretchen Krueger, and Ilya Sutskever.
\newblock {Learning Transferable Visual Models From Natural Language Supervision}.
\newblock In \emph{Proc.\ of ICML}, pages 8748--8763, virtual, 2021.

\bibitem[Russakovsky et~al.(2015)Russakovsky, Deng, Su, Krause, Satheesh, Ma, Huang, Karpathy, Khosla, Bernstein, Berg, and Fei-Fei]{Russakovsky2015}
Olga Russakovsky, Jia Deng, Hao Su, Jonathan Krause, Sanjeev Satheesh, Sean Ma, Zhiheng Huang, Andrej Karpathy, Aditya Khosla, Michael Bernstein, Alexander~C. Berg, and Li Fei-Fei.
\newblock {ImageNet Large Scale Visual Recognition Challenge}.
\newblock \emph{Int. J. Comput. Vis.}, 115\penalty0 (3):\penalty0 211--252, 2015.

\bibitem[Shazeer and Stern(2018)]{Shazeer2018}
Noam Shazeer and Mitchell Stern.
\newblock {Adafactor: Adaptive Learning Rates with Sublinear Memory Cost}.
\newblock In \emph{Proc.\ of ICML}, pages 4596--4604, Stockholm, Sweden, 2018.

\bibitem[Steiner et~al.(2022)Steiner, Kolesnikov, Zhai, Wightman, Uszkoreit, and Beyer]{Steiner2022}
Andreas Steiner, Alexander Kolesnikov, Xiaohua Zhai, Ross Wightman, Jakob Uszkoreit, and Lucas Beyer.
\newblock {How to Train Your ViT? Data, Augmentation, and Regularization in Vision Transformers}.
\newblock \emph{Trans. Mach. Learn. Res.}, pages 1--16, 2022.

\bibitem[Sun et~al.(2017)Sun, Shrivastava, Singh, and Gupta]{Sun2017}
Chen Sun, Abhinav Shrivastava, Saurabh Singh, and Abhinav Gupta.
\newblock {Revisiting Unreasonable Effectiveness of Data in Deep Learning Era}.
\newblock In \emph{Proc.\ of ICCV}, pages 843--852, Venice, Italy, 2017.

\bibitem[Szegedy et~al.(2016)Szegedy, Vanhoucke, Ioffe, Shlens, and Wojna]{Szegedy2016}
Christian Szegedy, Vincent Vanhoucke, Sergey Ioffe, Jon Shlens, and Zbigniew Wojna.
\newblock {Rethinking the Inception Architecture for Computer Vision}.
\newblock In \emph{Proc.\ of CVPR}, pages 2818--2826, Las Vegas, NV, USA, 2016.

\bibitem[Touvron et~al.(2021)Touvron, Cord, Douze, Massa, Sablayrolles, and Jegou]{Touvron2021}
Hugo Touvron, Matthieu Cord, Matthijs Douze, Francisco Massa, Alexandre Sablayrolles, and Herve Jegou.
\newblock {Training Data-Efficient Image Transformers \& Distillation Through Attention}.
\newblock In \emph{Proc.\ of ICML}, pages 10347--10357, virtual, 2021.

\bibitem[Verma et~al.(2019)Verma, Lamb, Beckham, Najafi, Mitliagkas, Lopez-Paz, and Bengio]{Verma2019}
Vikas Verma, Alex Lamb, Christopher Beckham, Amir Najafi, Ioannis Mitliagkas, David Lopez-Paz, and Yoshua Bengio.
\newblock {Manifold Mixup: Better Representations by Interpolating Hidden States}.
\newblock In \emph{Proc.\ of ICML}, pages 6438--6447, Long Beach, CA, USA, 2019.

\bibitem[Vinyals et~al.(2016)Vinyals, Blundell, Lillicrap, kavukcuoglu, and Wierstra]{Vinyals2016}
Oriol Vinyals, Charles Blundell, Timothy Lillicrap, koray kavukcuoglu, and Daan Wierstra.
\newblock {Matching Networks for One Shot Learning}.
\newblock In \emph{Proc.\ of NIPS}, pages 3637--3645, Barcelona, Spain, 2016.
\newblock (`NIPS' was renamed to `NeurIPS' after 2018).

\bibitem[Wang et~al.(2021)Wang, Xie, Li, Fan, Song, Liang, Lu, Luo, and Shao]{wang2021pyramid}
Wenhai Wang, Enze Xie, Xiang Li, Deng-Ping Fan, Kaitao Song, Ding Liang, Tong Lu, Ping Luo, and Ling Shao.
\newblock {Pyramid Vision Transformer: A Versatile Backbone for Dense Prediction without Convolutions}.
\newblock In \emph{Proc.\ of ICCV}, pages 548--558, virtual, 2021.

\bibitem[Xiao et~al.(2010)Xiao, Hays, Ehinger, Oliva, and Torralba]{Xiao2010}
Jianxiong Xiao, James Hays, Krista~A. Ehinger, Aude Oliva, and Antonio Torralba.
\newblock {SUN Database: Large-Scale Scene Recognition from Abbey to Zoo}.
\newblock In \emph{Proc.\ of CVPR}, pages 3485--3492, San Francisco, CA, USA, 2010.

\bibitem[Xiao et~al.(2016)Xiao, Ehinger, Hays, Torralba, and Oliva]{Xiao2016}
Jianxiong Xiao, Krista~A. Ehinger, James Hays, Antonio Torralba, and Aude Oliva.
\newblock {SUN Database: Exploring a Large Collection of Scene Categories}.
\newblock \emph{Int. J. Comput. Vis.}, 119\penalty0 (1):\penalty0 3--22, 2016.

\bibitem[Zhai et~al.(2020)Zhai, Puigcerver, Kolesnikov, Ruyssen, Riquelme, Lucic, Djolonga, Pinto, Neumann, Dosovitskiy, Beyer, Bachem, Tschannen, Michalski, Bousquet, Gelly, and Houlsby]{Zhai2020}
Xiaohua Zhai, Joan Puigcerver, Alexander Kolesnikov, Pierre Ruyssen, Carlos Riquelme, Mario Lucic, Josip Djolonga, Andr{\'{e}}~Susano Pinto, Maxim Neumann, Alexey Dosovitskiy, Lucas Beyer, Olivier Bachem, Michael Tschannen, Marcin Michalski, Olivier Bousquet, Sylvain Gelly, and Neil Houlsby.
\newblock {A Large-scale Study of Representation Learning with the Visual Task Adaptation Benchmark}.
\newblock \emph{arXiv}, 1910.04867:\penalty0 1--33, 2020.

\bibitem[Zhai et~al.(2022)Zhai, Kolesnikov, Houlsby, and Beyer]{Zhai2022}
Xiaohua Zhai, Alexander Kolesnikov, Neil Houlsby, and Lucas Beyer.
\newblock {Scaling Vision Transformers}.
\newblock In \emph{Proc.\ of CVPR}, pages 12104--12113, New Orleans, LA, USA, 2022.

\bibitem[Zhai et~al.(2023)Zhai, Mustafa, Kolesnikov, and Beyer]{Zhai2023}
Xiaohua Zhai, Basil Mustafa, Alexander Kolesnikov, and Lucas Beyer.
\newblock {Sigmoid Loss for Language Image Pre-Training}.
\newblock In \emph{Proc.\ of ICCV}, pages 11975--11986, Paris, France, 2023.

\bibitem[Zhang et~al.(2018)Zhang, Ciss{\'{e}}, Dauphin, and Lopez{-}Paz]{Zhang2018h}
Hongyi Zhang, Moustapha Ciss{\'{e}}, Yann~N. Dauphin, and David Lopez{-}Paz.
\newblock {Mixup: Beyond Empirical Risk Minimization}.
\newblock In \emph{Proc.\ of ICLR}, pages 1--13, Vancouver, BC, Canada, 2018.

\bibitem[Zhang et~al.(2022)Zhang, Zhang, Fang, Gao, Li, Dai, Qiao, and Li]{Zhang2022a}
Renrui Zhang, Wei Zhang, Rongyao Fang, Peng Gao, Kunchang Li, Jifeng Dai, Yu Qiao, and Hongsheng Li.
\newblock {Tip-Adapter: Training-Free Adaption of CLIP for Few-Shot Classification}.
\newblock In \emph{Proc.\ of ECCV}, pages 493--510, Tel Aviv, Israel, 2022.

\end{thebibliography}
}

\clearpage
\maketitlesupplementary

% Reset and redefine section numbering
\setcounter{section}{0}
\renewcommand{\thesection}{A\arabic{section}}
\renewcommand{\thesubsection}{A\arabic{section}.\arabic{subsection}}

\section{Introduction}
We give additional details and results to complement the main paper. All included citations refer to the main paper's references.

\section{Brightness}
We provide the code snippet for brightness $c^2$ FroFa

\begin{python}
def transform_aug_reverse(
  x, augment, aug_min_val=0, aug_max_val=1.0,
  x_min_val=None, x_max_val=None, clip=True):
  """Transform to (low, high)-space,
  perform augmentation, transform back."""
  l = x_min_val
  if x_min_val is None:
    l = tf.reduce_min(x)

  h = x_max_val
  if x_max_val is None:
    h = tf.reduce_max(x)

  # [l, h] --> [0, 1]
  x = (x - l) / (h - l + 1e-8)  

  # [0, 1] --> [low, high]
  x = x * (aug_max_val - aug_min_val)
  x = x + aug_min_val  
  x = tf.cast(augment(x), tf.float32)
  if clip:
    tf.clip_by_value(x, aug_min_val, aug_max_val)

  # [low, high] --> [0, 1]
  x = (x - aug_min_val)
  x = x / (aug_max_val - aug_min_val)  

  x = x * (h - l + 1e-8) + l  # [0, 1] --> [l, h]
  return x

def get_random_brightness(max_delta=0.1,
                          clip=False):
  # A random value in [-max_delta, +max_delta]
  # is added to the image values.
  # Small max_delta <1.0 assumes that the
  # image values are within [0, 1].
  def _random_brightness(image):
    return tf.image.random_brightness(
      image, max_delta)
  def tar(x):
    return transform_aug_reverse(
      x, augment=_random_brightness,
      aug_min_val=0, aug_max_val=1.0, clip=clip)
  return tar

def get_random_brightness_per_channel_v2(
    max_delta=0.1, clip=True):
  """Applies channel-wise random brightness
    transformations."""
  # A random value in [-max_delta, +max_delta] is
    added to the image values.
  # Small max_delta <1.0 assumes that the
  # image values are within [0, 1].
  random_brightness = get_random_brightness(
    max_delta, clip)
  def _random_brightness_pc(x):
    x = tf.expand_dims(x, axis=2)  # (H, W, 1, C)
    x = tf.unstack(x, axis=-1)  # C x (H, W, 1)
    x = [random_brightness(
      {"image": x_i})["image"] for x_i in x]
    return tf.concat(x, axis=-1)
  return _random_brightness_pc
\end{python}

\section{Detailed Experimental Setup}\label{sec:appendix_exp_setup}
In the following, we provide additional details to our experimental setup.

\subsection{Datasets}\label{sec:appendix_datasets}
In this section, we focus on details regarding our pretraining and few-shot datasets.

\textbf{Pretraining}: As stated in the main paper, \cref{sec:dataset_details}, we pretrain our models by either using JFT-3B \cite{Zhai2022}, ImageNet-21k \cite{Deng2009,Russakovsky2015}, or WebLI \cite{Chen2023}.

In JFT-3B, the images are annotated with noisy labels by using a semi-automated pipeline. We follow common practice \cite{Zhai2022,Dehghani2023} and ignore the hierarchical aspect of the labels. ImageNet-21k is a superset of the well known ILSVRC-2012 dataset, also known as ``ImageNet-1k'' or just ``ImageNet''. WebLI is a recently introduced image-and-language dataset. It contains 10 billion images and tens of billions image-text pairs with over 100 languages.

\textbf{Few-shot transfer}: As stated in the main paper, \cref{sec:dataset_details}, our experiments concentrate around few-shot transfer on ILSVRC-2012 \cite{Deng2009, Russakovsky2015}. We also provide results on
CIFAR10 \cite{Krizhevsky2009}, CIFAR100 \cite{Krizhevsky2009}, DMLab \cite{Beattie2016,Zhai2020}, DTD \cite{Cimpoi2014}, Resisc45 \cite{Cheng2017}, SUN397 \cite{Xiao2010,Xiao2016}, and SVHN \cite{Netzer2011}. When official test and validation splits are available, we use them for evaluation across all datasets. In general, we use the versions in TensorFlow Datasets\footnote{\url{https://www.tensorflow.org/datasets}}.

% ILSVRC-2012 \todo{[Introductory sentence]}. It contains 88,178 images of 6 classes. We use \todo{XXX} for training, \todo{XXX} for validation, and \todo{XXX} for test.

\textit{CIFAR10} contains 60,000 images of 10 equally distributed classes split into 50,000 training images and 10,000 test images. We further split the official training dataset into 45,000 training images and 5,000 validation images.

\textit{CIFAR100} is a superset of CIFAR10 with 100 equally distributed classes and 60,000 images. Similar to CIFAR10, we use 45,000 images for training, 5,000 images for validation and 10,000 images for test.

\textit{DMLab} consists of frames collected from the DeepMind Lab environment. Each frame is annotated with one out of six classes. We use 65,550 images for training, 22,628 images for validation, and 22,735 for test.

\textit{DTD} is a collection of 5,640 textural images categorized into 47 distinct classes. Each of the three splits, \ie, training, validation, and test, has exactly 1,880 images.

\textit{Resisc45} is a benchmark with 31,500 images for image scene classification in remote sensing scenarios. In total, 47 different catogries for scenes are defined. We use the first 23,000 images for training, the subsequent 2,000 images for validation and the last 6,300 images for test.

\textit{SUN397} is a 397-category database of 108,753 images for scene understanding. We use 76,128 images for training, 10,875 images for validation, and 21,750 images for test. 

\textit{SVHN} is a Google Street View dataset with a large collection of house number images. In total, 10 distinct classes exist. We use the cropped version with 73,257 images for training and 26,032 images for test. Further, we create a validation subset by only using the first 70,000 out of 73,257 training images for actual training and the remaining 3,257 images for validation. 

\subsection{Data Augmentation}\label{sec:appendix_augmentation}
\begin{table*}
\setlength{\tabcolsep}{2.8pt}
\small
  \centering
  \begin{tabular}{@{}clp{148mm}@{}}
    \toprule
    &Augmentation & Description \\
    \midrule
    \multirow{6}{*}{\rotatebox{90}{\footnotesize Geometric}}&rotate & We rotate each of the $C$ feature channels $\boldsymbol{f}_c$ \eqref{eq:single_feature} by $z\sim U(-v, v)$. We sweep across $v\in\lbrace15, 30, 45, 60, 75, 90\rbrace$ representing the maximum positive and negative rotation angle in degrees.\\
    & shear-$\lbrace$x,y$\rbrace$ & We (horizontally/vertically) shear each of the $C$ feature channels $\boldsymbol{f}_c$ \eqref{eq:single_feature} by $z\sim U(0, v)$. We sweep across $v\in\lbrace0.1, 0.2, 0.3, 0.4, 0.5, 0.6, 0.7\rbrace$ representing the maximum level of horizontal or vertical shearing.\\
    & translate-$\lbrace$x,y$\rbrace$ & We (horizontally/vertically) translate each of the $C$ feature channels $\boldsymbol{f}_c$ \eqref{eq:single_feature} by uniformly sampling $z$ from $\lbrace 0, 1, ..., v \rbrace$. We sweep across integer values $1 \le v \le 7$ representing the maximum horizontal or vertical translation.\\
    \midrule
    \multirow{7}{*}{\rotatebox{90}{\footnotesize Crop \& drop}}& crop& We randomly crop each of the $C$ feature channels $\boldsymbol{f}_c$ \eqref{eq:single_feature} to $v\times v$ at the same spatial position. We sweep across integer values $1 \le v \le 13$ representing the square crop size.\\
    & resized crop& We resize each of the $C$ feature channels $\boldsymbol{f}_c$ \eqref{eq:single_feature} to $v\times v$ and then randomly crop each to $14\times 14$ at the same spatial position. We sweep across $v\in\lbrace 16, 18, 20, 22, 24, 26, 28, 35, 42\rbrace$ representing the resized squared spatial resolution.\\
    & inception\ crop& We apply an inception crop with probability $v$. We sweep across $v\in\lbrace 0.1, 0.2, 0.3, 0.4, 0.5, 0.6, 0.7, 0.8, 0.9, 1.0 \rbrace$.\\
    & patch dropout\ & We randomly keep $v$ out of $N$ patches of $\boldsymbol{f}$ having shape $N\times C$. Note that the patch ordering is also randomized. We sweep across $v\in\lbrace 1, 2, 4, 12, 20, 28, 36, 44, 52, 60, 68, 76, 84, 92, 100, 116, 132, 148, 164, 180 \rbrace$.\\
    \midrule
    \multirow{26}{*}{\rotatebox{90}{\footnotesize Stylistic}}& brightness & We randomly add a value $z\sim U(-v, v)$ to each of the $C$ feature channels $\boldsymbol{f}_c$ \eqref{eq:single_feature}. We sweep across $v\in\lbrace 0.1, 0.2, 0.3, 0.4, 0.5, 0.6, 0.7, 0.8, 0.9, 1.0 \rbrace$. We test this method using all FroFA variants. In the default FroFA and the cFroFA variants, the features are scaled by \eqref{eq:f_to_x_specific} taking the minimum $f_\mathrm{min}$ and maximum $f_\mathrm{max}$ across all channels into account. In the c$^2$FroFA variant, each channel $\boldsymbol{f}_c$ \eqref{eq:single_feature} is shifted individually and uses the channel minimum and maximum instead. Further, in the cFroFA and c$^2$FroFA variants we sample $C$ values of $z$, one for each channel.\\
    & contrast & We randomly scale each of the $C$ feature channels $\boldsymbol{f}_c$ \eqref{eq:single_feature} by $z\sim U(\frac{1}{v}, v)$. We sweep across $v\in\lbrace 1.25, 1.5, 2, 3, 4, 5, 6, 7, 9, 10 \rbrace$. We test this method using the default FroFA as well as cFroFA. Note that in the cFroFA variant we sample $C$ values of $z$, one for each channel.\\
    & equalize & We first map the features from value range $\mathbb{R}$ to the integer subset $\mathbb{I}=\lbrace 0, 1, ... , 195\rbrace$, \ie, executing \eqref{eq:f_to_x_specific} followed up by a discretization step. We choose this value range as preliminary results mapping from $\mathbb{R}$ to the more commonly used $\mathbb{I}=\lbrace 0, 1, ... , 255\rbrace$ instead didn't show any effects. We continue by equalizing 196 bins and then transforming the results back to the original space using \eqref{eq:x_to_f_specific}. We apply equalize with probability $v$. In particular, we sweep across $v\in\lbrace 0.1, 0.2, 0.3, 0.4, 0.5, 0.6, 0.7, 0.8, 0.9 \rbrace$. \\
    & invert& We change the sign of features $\boldsymbol{f}^*$ with probability $v$. We sweep across $v\in\lbrace 0.1, 0.2, 0.3, 0.4, 0.5, 0.6, 0.7, 0.8, 0.9 \rbrace$.\\
    & posterize& We first map the features from value range $\mathbb{R}$ to the integer subset $\mathbb{I}=\lbrace 0, 1, ... , 255\rbrace$, \ie, executing \eqref{eq:f_to_x_specific} followed up by a discretization step. In other words, we use an 8-bit representation for features $\boldsymbol{f}^*$. Posterize performs a quantization by a bit-wise left and right shift. We uniformly sample the shift value $z$ between integer values $v_1$ and $v_2$. In our sweep, we test a subset of all possible combinations. In particular, we first set $v_2=8$ and reduce $v_1$ from 7 to 1. We then fix $v_1=1$ and increase $v_2$ from 2 to 7 again. We test this method using the default FroFA as well as cFroFA. Note that in the cFroFA variant we sample $C$ values of $z$, one for each channel.\\
    & sharpness& We first apply a two-dimensional convolution on $\boldsymbol{f}^*$ \eqref{eq:reshaped_features} using a $3\!\times\!3$ smoothing filter. Next, we mix the original features with the resulting ``smoothed'' features using a randomly sampled blending factor $z\sim U(0, v)$. We sweep across $v\in\lbrace 0.2, 0.4, 0.6, 0.8, 1.0, 1.5, 2.0, 3.0 \rbrace$.\\
    & solarize& We do not map features from $\mathbb{R}$ to $\mathbb{I}=[0,1]$, but stay in $\mathbb{R}$. We compute the minimum $f_\mathrm{min}$ and maximum $f_\mathrm{max}$ across features $\boldsymbol{f}^*$. We conditionally subtract all values smaller than $0.5\cdot f_\mathrm{min}$ from $f_\mathrm{min}$ or larger than $0.5\cdot f_\mathrm{max}$ from $f_\mathrm{max}$. We apply this method with a probability $v$ and sweep across $v\in\lbrace 0.1, 0.2, 0.3, 0.4, 0.5, 0.6, 0.7, 0.8, 0.9, 1.0 \rbrace$.\\
    \midrule
    \multirow{6}{*}{\rotatebox{90}{\footnotesize Other}}& JPEG& We first map the features from value range $\mathbb{R}$ to the integer subset $\mathbb{I}=\lbrace 0, 1, ... , 255\rbrace$, \ie, executing \eqref{eq:f_to_x_specific} followed up by a discretization step. We then perform a JPEG compression of each channel by randomly sampling a JPEG quality $z\sim U(v_1, v_2)$. We sweep across combinations of $v_1\in\lbrace10, 25, 50, 75\rbrace$ and $v_2\in\lbrace25, 50, 75,100\rbrace$, with $v_2 > v_1$.\\
    & mixup & We do not map features from $\mathbb{R}$ to $[0,1]$, but stay in $\mathbb{R}$. We mix two features $\boldsymbol{f}^*_i, \boldsymbol{f}^*_j$ according to $z\cdot\boldsymbol{f}^*_i + (1-z) \cdot \boldsymbol{f}^*_j$ by sampling a random value $z\sim B(\alpha,\alpha)$, with Beta distribution $B(\alpha,\alpha)$ parameterized by $\alpha=v$. The labels are mixed using the same procedure. We sweep across $v\in\lbrace 0.025, 0.05, 0.1, 0.2, 0.3, 0.4, 0.5, 0.6, 0.7, 0.8, 0.9, 1.0 \rbrace$.\\
    \bottomrule
  \end{tabular}
  \caption{\textbf{Details on our used set of augmentations}. For simplicity, instead of introducing a new hyper parameter for each data augmentation, we re-use $v$ as a sweep parameter that is set during a sweep and differs for each augmentation. If not stated otherwise, each method is only applied as default FroFA and we first map features $\boldsymbol{f}$ (two-dimensional representation) or $\boldsymbol{f}^*$ (three-dimensional representation) from value range $\mathbb{R}$ to $\mathbb{I}=[0,1]$ using \eqref{eq:f_to_x_specific}. By default, we assume a three-dimensional representation $\boldsymbol{f}^*$ although some augmentations would work also in the two-dimensional representation $\boldsymbol{f}$, \ie, a reshaping is not necessary.}
  \label{tab:appendix_augmentation_details}
\end{table*}
In this section, we provide additional details on the used data augmentation techniques and protocols.

\textbf{(c/c$^2$)FroFA}: In \cref{tab:appendix_augmentation_details}, we give detailed descriptions of each FroFA, cFroFA, and c$^2$FroFA setting. We mostly build upon an AutoAugment implementation from \texttt{Big Vision}\footnote{\url{https://github.com/google-research/big_vision/blob/main/big_vision/pp/autoaugment.py}}. To keep it simple, we use $v$ or $v_1,v_2$ as sweep parameter(s) for all augmentations. By default, we first reshape the two-dimensional features $\boldsymbol{f}$ to three-dimensional features $\boldsymbol{f}^*$ \eqref{eq:reshaped_features} of shape $\sqrt{N}\times\sqrt{N}\times C$, with $N=196$ and $C\in\lbrace 192, 768, 1024 \rbrace$ in all our experiments. Note that the value of $C$ depends on the architecture. We further want to point out, while some augmentations heavily rely on the three-dimensional representation, \eg, all geometric ones, some others are also transferable to a two-dimensional representation, \eg, brightness or contrast.

As pointed out in the main paper, \cref{tab:feature-wise}, brightness c$^2$FroFA, contrast FroFA, and posterize cFroFA are our best FroFAs. For all three, we list the best sweep settings in \cref{tab:appendix_augmentation_details_top3}.

\begin{table}
\setlength{\tabcolsep}{2.6pt}
\footnotesize
  \centering
  \begin{tabular}{@{}lrrrrc@{}}
    \toprule
    FroFA &Shots & Base learning rate & Batch size & Training steps & $v$ or $v_1,v_2$\\
    \midrule
    \multirow{4}{*}{B-c$^2$} &1 & 0.01 & 512 & 4,000 & 1.0\\
    &10 & 0.01 & 64 & 16,000 & 1.0\\
    &15 & 0.01 & 256 & 8,000 & 0.9\\
    &25 & 0.01 & 512 & 8,000 & 0.8\\
    \midrule
    \multirow{4}{*}{C} &1 & 0.01 & 32 & 16,000 & 6.0\\
    &10 & 0.01 & 128 & 8,000 & 6.0\\
    &15 & 0.01 & 512 & 2,000 & 6.0\\
    &25 & 0.01 & 256 & 4,000 & 7.0\\
    \midrule
    \multirow{4}{*}{P-c} &1 & 0.01 & 512 & 8,000 & 1, 8\\
    &10 & 0.03 & 512 & 8,000 & 1, 8\\
    &15 & 0.03 & 512 & 16,000 & 1, 8\\
    &25 & 0.03 & 64 & 16,000 & 2, 8\\
    \bottomrule
  \end{tabular}
  \caption{\textbf{Our best sweep settings for our best three FroFAs}, namely, brightness cFroFA (B-c$^2$), contrast (C), and posterize cFroFA (P-c). We list the shots, base learning rate, batch size, number of training steps, and the augmentation parameter, denoted as $v$ or $v_1,v_2$ (see \cref{tab:appendix_augmentation_details} for a detailed explanation of $v$ and $v_1, v_2$). The best sweep settings are found using \textit{our} ILSVRC-2012 validation set.}
  \label{tab:appendix_augmentation_details_top3}
\end{table}

\textbf{Advanced protocols}: As mentioned in the main paper, \cref{sec:augmentation_details}, besides our fixed sequential protocol (\cf \cref{tab:sequential}) we also tested variations of RandAugment \cite{Cubuk2020} and TrivialAugment \cite{Mueller2021}. In all protocols, we sample from the best settings of brightness c$^2$FroFA, contrast FroFA, and posterize cFroFA. In particular, we use $v=1.0$ for brightness c$^2$FroFA, $v=6.0$ for contrast FroFA, and $v_1=1, v_2=8$ for posterize cFroFA (\cf \cref{tab:appendix_augmentation_details}). We re-use the abbreviations from \cref{tab:sequential} in the following, \ie, B-c$^2$, C, and P-c, respectively. For the RandAugment and TrivialAugment variations, we uniformly sample from either the best three FroFAs, \ie, $\mathcal{A}_{\mathrm{top}3}=\lbrace \text{B-c}^2, \text{C}, \text{P-c} \rbrace$, or the best two FroFAs, \ie, $\mathcal{A}_{\mathrm{top}2}=\mathcal{A}_3\setminus\lbrace \text{C} \rbrace$. Further, our RandAugment variation randomly constructs a sequence of augmentations by uniformly sampling the integer sequence length from 1 to $|\mathcal{A}|$, with $\mathcal{A}\in\lbrace \mathcal{A}_{\mathrm{top}2}, \mathcal{A}_{\mathrm{top}3} \rbrace$ depending on whether $\mathcal{A}_{\mathrm{top}2}$ or $\mathcal{A}_{\mathrm{top}3}$ is used.

\subsection{Training Details}\label{sec:appendix_training_details}
\textbf{Pretraining}: In the JFT-3B setup, we use pretrained models from Zhai \etal \cite{Zhai2022}. The models are pretrained using a sigmoid cross-entropy loss. The weights are optimized by Adafactor \cite{Shazeer2018}
in half-precision mode, $\beta_1=0.9$, and $\beta_2=0.999$. Further, (decoupled) weight decay \cite{Loshchilov2019} is applied with 3.0 on the head and 0.03 for the rest of the network weights. The learning rate is adapted by a reciprocal square-root schedule for 4,000,000 steps with a linear warm-up phase of 10,000 steps and a linear cool-down phase of 50,000 steps. The starting learning rate is 0.01 for Ti/16 and L/16 and 0.03 for B/16. The images are preprocessed by an $224\!\times\!224$ inception crop and a random horizontal flip. We set the batch size to 4,096. To stabilize training, a global norm clipping of 1.0 is used.

In the ImageNet-21k setup, we follow settings from Steiner \etal \cite{Steiner2022} and use a sigmoid cross-entropy loss for multi-label pretraining. We use the Adam optimizer \cite{Kingma2015}
in half-precision mode and set $\beta_1=0.9$ and $\beta_2=0.999$. Further, we apply (decoupled) weight decay with either 0.03 for Ti/16 or 0.1 for B/16 and L/16. We adapt the learning rate using a cosine schedule for roughly 930,000 steps (300 epochs) with a linear warm-up phase of 10,000 steps. We set the starting learning rate to 0.001 for all models. During preprocessing, we crop the images to $224\!\times\!224$ following an inception-style crop and a random horizontal flip. While we don't use any additional augmentation for Ti/16, we follow suggestions by Steiner \etal \cite{Steiner2022} and use the `light1' and `medium2' augmentation settings for B/16 and L/16, respectively. Finally, we use a batch size of 4,096 and stabilize training by using a global norm clipping of 1.0.

In the WebLI setup, we take an L/16 model from \cite{Zhai2023}. In particular, we use \todo{[ADD DETAILS]}.

\textbf{Few-shot learning}: We first cache each few-shot dataset by processing each of them through a pretrained model and store the extracted features (\cf \cref{fig:framework_overview}). We resize each image to $224\!\times\!224$ before feeding it to the model.

We follow up with a training where we mostly use transfer learning settings from Steiner \etal \cite{Steiner2022}. We use a sigmoid cross-entropy loss. This might be non-intuitive given that all of our few-shot datasets are not multi-labeled. However, we didn't really observe any performance drops compared to using the more common softmax cross-entropy loss, so we stick to the sigmoid cross-entropy loss. We use stochastic gradient descent with momentum of 0.9. Similar to the pretraining setup, we also store the internal state in half-precision. We do not apply any weight decay. The learning rate is adapted following a cosine schedule with a linear warm-up phase of 500 steps. In addition, we stabilize training by using a global norm clipping of 1.0. Further, we sweep across batch size, learning rate and number of steps yielding 100 combinations (\cf \cref{sec:training_details}) for each shot.

\section{Additional Experimental Results}\label{sec:appendix_exp_results}
In this section, we show additional experimental results.

\subsection{Patch Dropout and Brightness}\label{sec:appendix_exp_patch_drop_and_brightness}
\begin{figure*}
  \centering
\includegraphics[width=\textwidth]{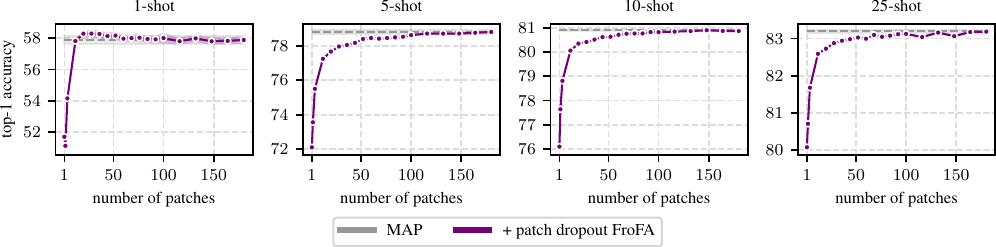}
  \caption{\textbf{Average top-1 accuracy for patch dropout FroFA} on \textit{our} ILSVRC-2012 test set. We use the L/16 JFT-3B base setup (\cf \cref{sec:exp_results}). We sweep across a base sweep (\cf \cref{sec:training_details}) to first find the best setting on \textit{our} ILSVRC-2012 validation set for each number of patches (\cf \cref{sec:appendix_augmentation}). Shaded areas indicate standard errors collected via sampling each shot five times.}  \label{fig:patch_dropout_all}
\end{figure*}

\begin{figure*}
  \centering
\includegraphics[width=\textwidth]{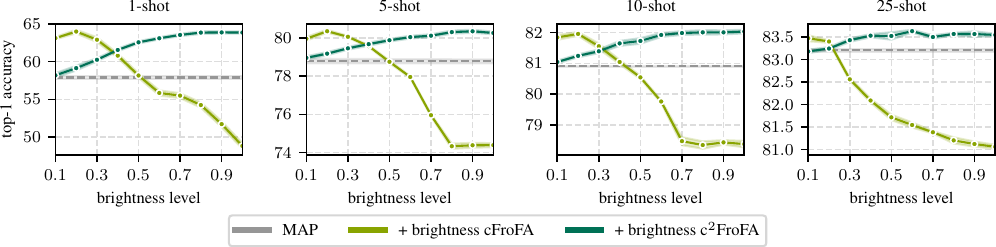}
  % \caption{\textbf{Top-1 accuracy for patch dropout FroFA} on \textit{our} ILSVRC-2012 test set. We use an L/16 pretrained on JFT-3B and transferred to few-shot subsets of the ILSVRC-2012 training set. We sweep across a base sweep (\cf \cref{sec:training_details}) to find the best setting on \textit{our} ILSVRC-2012 validation set for each \#patches, \ie, number of patches (\cf Appendix, \cref{sec:appendix_augmentation}). Shaded areas indicate standard errors collected via sampling each shot five times.}  \label{fig:patch_dropout}
  \caption{\textbf{Top-1 accuracy for channel variants (c/c$^2$) of brightness FroFA} on \textit{our} ILSVRC-2012 test set. We use the L/16 JFT-3B base setup (\cf \cref{sec:exp_results}). We sweep across a base sweep (\cf \cref{sec:training_details}) to first find the best setting on \textit{our} ILSVRC-2012 validation set for each brightness level (\cf \cref{sec:appendix_augmentation}). Shaded areas indicate standard errors collected via sampling each shot five times}  \label{fig:brightness-comparison_all}
\end{figure*}

In \cref{fig:brightness+pd-comparison}, we only report results for 1-shot and 25-shot settings using patch dropout FroFA and brightness (c/c$^2$)FroFA. We extend this by also reporting results for 5-shot and 10-shot settings in \cref{fig:brightness-comparison_all,fig:patch_dropout_all}. We observe the same effects in the other settings as well. 

\begin{table}
\setlength{\tabcolsep}{2.8pt}
\footnotesize
  \centering
  \begin{tabular}{@{}lccccccccc@{}}
    \toprule
    & && && \multicolumn{2}{c}{RA$^*$} && \multicolumn{2}{c}{TA$^*$}\\
    Shots & MAP && \rotatebox{0}{{B-c$^2$}}&& $\mathcal{A}_{\mathrm{top}2}$ & $\mathcal{A}_{\mathrm{top}3}$ && $\mathcal{A}_{\mathrm{top}2}$ & $\mathcal{A}_{\mathrm{top}3}$\\
    \midrule
    1 &  58.4&&\textbf{$+$6.0}&&$+$3.9&$+$2.4&&$+$4.8&$+$4.3\\
    5 &  79.1 && \textbf{$+$1.5}&&$+$1.0&$+$0.4&&\textbf{$+$1.4}&$+$1.2\\
    10 &  80.7 && \textbf{$+$1.3}&&$+$1.0&$+$0.6&&\textbf{$+$1.4}&$+$1.4\\
    25 &  83.0 &&\textbf{$+$0.6}&&\textbf{$+$0.4}&$+$0.0&&\textbf{$+$0.5}&$+$0.4\\
    \bottomrule
  \end{tabular}
  \caption{\textbf{Top-1 accuracy for advanced FroFA protocols} on \textit{our} ILSVRC-2012 test set. Absolute gains to the MAP baseline (reference run) are reported. We use the L/16 JFT-3B base setup (\cf \cref{sec:exp_results}). We compare brightness c$^2$FroFA (B-c$^2$) with our variations of RandAugment (RA$^*$) and TrivialAugment (TA$^*$), \cf \cref{sec:appendix_augmentation}. For the latter, we either use the top-2 ($\mathcal{A}_{\mathrm{top}2}$) or top-3 ($\mathcal{A}_{\mathrm{top}3}$) augmentations. We sweep across a base sweep (\cf \cref{sec:training_details}) to first find the best setting on \textit{our} ILSVRC-2012 validation set. The best results per shot are boldfaced (multiple ones if close, \ie, $\pm$0.2).}
  \label{tab:appendix_advanced}
\end{table}

\subsection{Advanced FroFA Protocols}\label{sec:appendix_exp_advanced_protocols}
In \cref{tab:appendix_advanced}, we report results for our RandAugment (RA$^*$) and TrivialAugment (TA$^*$) variations. We did not average across five runs and thus only report absolute gains with respect to a reference run. Therefore, numbers which are also reported in the main paper, \eg, \cref{tab:sequential}, are slightly different. All in all, we observe that both RA$^*$ and TA$^*$ do not improve upon the best single augmentation, \ie, brightness c$^2$FroFA (B-c$^2$). We also observe that increasing the set of augmentations from $\mathcal{A}_{\mathrm{top}2}$ to $\mathcal{A}_{\mathrm{top}3}$ rather worsens the performance for both RA$^*$ and TA$^*$.

\subsection{Detailed FroFA Transfer Results}\label{sec:appendix_exp_frofa_transfer}
\begin{table*}
\setlength{\tabcolsep}{2.8pt}
\footnotesize
  \centering
  \begin{tabular}{@{}llcccccccccc@{}}
    \toprule
    & && \multicolumn{4}{c}{ImageNet-21k} && \multicolumn{4}{c}{JFT-3B}\\
    Model & Method && 1-shot & 5-shot& 10-shot & 25-shot && 1-shot & 5-shot& 10-shot & 25-shot\\
    \midrule
    \multirow{3}{*}{Ti/16} & MAP\textsuperscript{wd} &&20.5&53.6&59.7&64.9&&19.1&46.4&\textbf{53.6}&\textbf{60.2}\\
    & Linear probe &&\textbf{36.8}&53.7&58.0&61.1&&\textbf{33.0}&\textbf{48.0}&52.2&55.4\\
    &MAP\textsuperscript{wd} + FroFA &&20.6&\textbf{54.5}&\textbf{60.1}&\textbf{65.2}&&19.6&47.2&\textbf{53.6}&\textbf{60.3}\\
    \midrule
    \multirow{3}{*}{B/16} & MAP\textsuperscript{wd}&&30.5&71.7&75.3&\textbf{78.0}&&51.3&74.8&\textbf{77.5}&\textbf{79.8}\\
    & Linear probe&&\textbf{52.2}&72.9&\textbf{76.0}&\textbf{77.9}&&\textbf{59.6}&74.5&76.9&78.3\\
    &MAP\textsuperscript{wd} + FroFA&&30.6&\textbf{73.3}&\textbf{76.0}&\textbf{78.1}&&52.5&\textbf{75.1}&\textbf{77.6}&79.5\\
    \midrule
    \multirow{3}{*}{L/16} & MAP\textsuperscript{wd}&&38.7&75.9&78.6&80.6&&62.0&79.9&81.5&83.2\\
    & Linear probe&&\textbf{54.7}&77.1&\textbf{79.8}&\textbf{81.1}&&\textbf{66.5}&79.6&81.5&82.4\\
    &MAP\textsuperscript{wd} + FroFA&&39.3&\textbf{78.0}&\textbf{80.0}&\textbf{81.0}&&63.7&\textbf{80.4}&\textbf{82.0}&\textbf{83.6}\\
    \bottomrule
  \end{tabular}
  \caption{\textbf{Average top-1 accuracy for JFT-3B and ImageNet-21k models} on \textit{our} ILSVRC-2012 test set trained on few-shotted ILSVRC-2012 training sets. We report results for the weight-decayed MAP, \ie MAP\textsuperscript{wd}, and L2-regularized linear probe baseline, as well as our best FroFA-based approach, \ie, weight-decayed MAP combined with brightness c$^2$FroFA (MAP\textsuperscript{wd} + FroFA). Depending on the setting, we sweep across a base, \cf \cref{sec:training_details}, a weight decay or L2 decay, \cf \cref{sec:baselines}, and a brightness level sweep, \cf \cref{sec:appendix_augmentation}, to first find the best setting on \textit{our} ILSVRC-2012 validation set for each model. The best results per shot are boldfaced (multiple ones if close, \ie, $\pm$0.2). Our approach,  \ie, MAP\textsuperscript{wd} + FroFA, is on par or significantly better than MAP\textsuperscript{wd} and/or linear probe on most 5- to 25-shot settings.}
  \label{tab:appendix_i21k_jft3b_wd_all}
\end{table*}
\begin{table*}
\setlength{\tabcolsep}{2.8pt}
\footnotesize
  \centering
  \begin{tabular}{@{}llcccccccccc@{}}
    \toprule
    & && \multicolumn{4}{c}{ImageNet-21k} && \multicolumn{4}{c}{JFT-3B}\\
    Model & Method && 1-shot & 5-shot& 10-shot & 25-shot && 1-shot & 5-shot& 10-shot & 25-shot\\
    \midrule
    \multirow{3}{*}{Ti/16} & MAP &&20.4&53.2&59.5&64.7&&17.9&45.5&\textbf{53.5}&\textbf{60.1}\\
    & Linear probe &&\textbf{36.8}&53.7&58.0&61.1&&\textbf{33.0}&\textbf{48.0}&52.2&55.4\\
    &MAP + FroFA &&22.1&\textbf{54.9}&\textbf{60.1}&\textbf{65.0}&&20.3&47.2&\textbf{53.6}&\textbf{60.1}\\
    \midrule
    \multirow{3}{*}{B/16} & MAP&&31.3&70.3&75.1&\textbf{78.1}&&48.9&73.4&76.5&79.4\\
    & Linear probe&&\textbf{52.2}&72.9&76.0&77.9&&\textbf{59.6}&74.5&76.9&78.3\\
    &MAP + FroFA&&30.6&\textbf{73.4}&\textbf{76.3}&\textbf{78.3}&&52.4&\textbf{75.2}&\textbf{77.8}&\textbf{79.9}\\
    \midrule
    \multirow{3}{*}{L/16} & MAP&&38.8&74.9&78.5&80.7&&57.9&78.8&80.9&83.2\\
    & Linear probe&&\textbf{54.7}&77.1&\textbf{79.8}&\textbf{81.1}&&\textbf{66.5}&79.6&81.5&82.4\\
    &MAP + FroFA&&39.3&\textbf{78.0}&\textbf{80.0}&\textbf{81.2}&&63.9&\textbf{80.3}&\textbf{82.0}&\textbf{83.6}\\
    \bottomrule
  \end{tabular}
  \caption{\textbf{Average top-1 accuracy for JFT-3B and ImageNet-21k models} on \textit{our} ILSVRC-2012 test set trained on few-shotted ILSVRC-2012 training sets. We report results for the MAP and L2-regularized linear probe baseline, as well as our best FroFA-based approach, \ie, MAP combined with brightness c$^2$FroFA (MAP + FroFA). Depending on the setting, we sweep across a base, \cf \cref{sec:training_details}, an L2 decay, \cf \cref{sec:baselines}, and a brightness level sweep, \cf \cref{sec:appendix_augmentation}, to first find the best setting on \textit{our} ILSVRC-2012 validation set for each model. The best results per shot are boldfaced (multiple ones if close, \ie, $\pm$0.2). Our approach,  \ie, MAP + FroFA, is on par or significantly better than MAP and linear probe on most 5- to 25-shot settings.}
  \label{tab:appendix_i21k_jft3b_all}
\end{table*}
In \cref{tab:appendix_i21k_jft3b_wd_all}, we report exact numbers for \cref{fig:jft3b_and_i21k_all}, \ie, Ti/16, B/16, and L/16 pretrained on either ImageNet-21k or JFT-3B and subsequently finetuned on few-shotted ILSVRC-2012 training sets. Numbers for the two baselines, \ie, MAP (\textit{with} weight decay) and linear probe, and our best method, \ie, MAP (\textit{with} weight decay) combined with brightness c$^2$FroFA (MAP + FroFA), are reported. In addition, we report numbers, where we use MAP \textit{without} weight decay in \cref{tab:appendix_i21k_jft3b_all}. As before, we observe that our method performs worse on all 1-shot settings, but is on par or significantly better than MAP and/or linear probe on most 5- to 25-shot settings.
% {
%     \small
%     \bibliographystylesupp{ieeenat_fullname}
%     \bibliographysupp{supp}
% }

\end{document}